%% file: 8981.tex
\documentclass[10pt,twocolumn,letterpaper]{article}

\usepackage{iccv}
\usepackage{times}
\usepackage{epsfig}
\usepackage{graphicx}
\usepackage{amsmath}
\usepackage{amssymb}

\usepackage[accsupp]{axessibility}  % Improves PDF readability for those with disabilities.

\input{Preamble/preamble.tex}
\input{Preamble/acronyms.tex}

\input{Preamble/symbols.tex}

\def\codelink{\url{https://github.com/jspenmar/slowtv_monodepth}}
\graphicspath{{Images/}}
\usepackage[pagebackref=true,breaklinks=true,letterpaper=true,colorlinks,bookmarks=false]{hyperref}

\iccvfinalcopy % *** Uncomment this line for the final submission

\def\iccvPaperID{8981} % *** Enter the ICCV Paper ID here

\begin{document}

%%%%%%%%% TITLE
\title{Kick Back \& Relax: Learning to Reconstruct the World by Watching SlowTV}

\author{%
Jaime Spencer\\
University of Surrey\\
\parbox{5.5cm}{\centering\tt\small j.spencermartin@surrey.ac.uk}
\and
Chris Russell\\
Oxford Internet Institute\\
\parbox{5.5cm}{\centering\tt\small christopher.m.russell@gmail.com}
\\\and
Simon Hadfield\\
University of Surrey\\
\parbox{5.5cm}{\centering\tt\small s.hadfield@surrey.ac.uk}
\and
Richard Bowden\\
University of Surrey\\
\parbox{5.5cm}{\centering\tt\small r.bowden@surrey.ac.uk}
}

% \maketitle
\twocolumn[{%
\renewcommand\twocolumn[1][]{#1}%
\maketitle
\centering
\captionsetup{type=figure}
\captionsetup[subfigure]{labelformat=empty}
\input{Figures/intro}
\label{fig:intro}
\vspace{2ex}
}]


%%%%%%%%% ABSTRACT
\begin{abstract}
\vspace{-0.2cm}
Self-supervised monocular depth estimation (SS-MDE) has the potential to scale to vast quantities of data. Unfortunately, existing approaches limit themselves to the automotive domain, resulting in models incapable of generalizing to complex environments such as natural or indoor settings.

To address this, we propose a large-scale SlowTV dataset curated from YouTube, containing an order of magnitude more data than existing automotive datasets. SlowTV contains 1.7M images from a rich diversity of environments, such as worldwide seasonal hiking, scenic driving and scuba diving. Using this dataset, we train an SS-MDE model that provides zero-shot generalization to a large collection of indoor/outdoor datasets. The resulting model outperforms all existing SSL approaches and closes the gap on supervised SoTA, despite using a more efficient architecture. 

We additionally introduce a collection of best-practices to further maximize performance and zero-shot generalization. This includes 1) aspect ratio augmentation, 2) camera intrinsic estimation, 3) support frame randomization and 4) flexible motion estimation. Code is available at {\footnotesize \codelink}.
\end{abstract}

%%%%%%%%% BODY TEXT
%-------------------------------------------------------------------------
\section{Introduction} \label{sec:intro}
Reliably reconstructing the \ndim{3} structure of the environment is a crucial component of many computer vision pipelines, including autonomous driving, robotics, augmented reality and scene understanding. 
Despite being an inherently ill-posed task, \ac{mde} has become of great interest due to its flexibility and applicability to many fields.

While traditional supervised methods achieve impressive results, they are limited both by the availability and quality of annotated datasets. 
\acs{lidar} data is expensive to collect and frequently exhibits boundary artefacts due to motion correction. 
Meanwhile, \ac{sfm} is computationally expensive and can produce noisy, incomplete or incorrect reconstructions.
\Ac{ssl} instead leverages the photometric consistency across frames to simultaneously learn depth and \ac{vo} without ground-truth annotations. 
As only stereo or monocular video is required, \ac{ssl} has the potential to scale to much larger data quantities. 

Unfortunately, existing \acl{ssmde} approaches have relied exclusively on automotive data~\cite{Geiger2013,Cordts2016,Guizilini2020}. 
The limited diversity of training environments results in models incapable of generalizing to different scene types (\eg natural or indoors) or even other automotive datasets. 
Moreover, despite being fully convolutional, these models struggle to adapt to different image sizes.
This further reduces performance on sources other than the original dataset. 

Inspired by the recent success of supervised \ac{mde}~\cite{Xuan2020,Ranftl2020,Ranftl2021}, we develop an \acl{ssmde} model capable of performing zero-shot generalization beyond the automotive domain.
In doing so, we aim to bridge the performance gap between supervision and self-supervision.
Unfortunately, most existing supervised datasets are unsuitable for \ac{ssl}, as they consist of isolated image and depth pairs.
On the other hand, existing \ac{ssl} datasets focus only on the automotive domain.

To overcome this, we make use of \acl{stv} as an untapped source of high-quality data. 
\acl{stv} is a television programming approach originating from Norway consisting of long, uninterrupted shots of relaxing events, such as train or boat journeys, nature hikes and driving. 
This represents an ideal training source for \acl{ssmde}, as it provides large quantities of data from highly diverse environments, usually with smooth motion and limited dynamic objects. 

To improve the diversity of available data for \acl{ssmde}, we have collated the \textbf{\acl{stv}} dataset, consisting of 1.7M frames from 40 videos curated from YouTube. 
This dataset consists of three main categories---natural, driving and underwater---each featuring a rich and diverse set of scenes. 
 We combine \acl{stv} with \acl{mc}~\cite{Li2020} and Kitti~\cite{Geiger2013} to train our proposed models.
\acl{stv} provides a general distribution across a wide range of natural scenes, while \acl{mc} covers indoor scenes with humans and Kitti focuses on urban scenes. 
The resulting models are trained with an order of magnitude more data than any existing \acl{ssmde} approach.
Contrary to many supervised approaches~\cite{Bhat2021,Weihao2022}, we train a single model capable of generalizing to all scene types,
rather than separate indoor/outdoor models. 
This closely resembles the zero-shot evaluation proposed by \acl{midas}~\cite{Ranftl2020} for supervised MDE.

The contributions of this paper can be summarized as:
% =====================================
\begin{enumerate}
    \item We introduce a novel \acl{ssmde} dataset of SlowTV YouTube videos, consisting of 1.7M images.
    It features a diverse range of environments including worldwide seasonal hiking, scenic driving and scuba diving. 
    \vspace{-0.5cm}
    \item We leverage \acl{stv} to train zero-shot models capable of adapting to a wide range of scenes. 
    The models are evaluated on 7 datasets unseen during training. 
    \vspace{-0.2cm}
    \item We show that existing models fail to generalize to different image shapes and propose an aspect ratio augmentation to mitigate this.
    \vspace{-0.2cm}
    \item We greatly reduce the performance gap \wrt supervised models, improving the applicability of \acl{ssmde} to the real-world.
    We make the dataset, pretrained model and code available to the public. 
\end{enumerate}
% =====================================

%------------------------------------------------------------------------
\section{Related Work} \label{sec:lit}
Garg~\etal~\cite{Garg2016} proposed the first algorithm for \acl{ssmde}, where the target view was synthesized using its stereo pair and predicted depth map. 
Monodepth~\cite{Godard2017} greatly improved performance by incorporating differentiable bilinear interpolation~\cite{Jaderberg2015}, an SSIM-weighted reconstruction loss~\cite{Wang2004} and left-right consistency. 
SfM-Learner~\cite{Zhou2017} extended \acl{ssmde} into the purely monocular domain by replacing the fixed stereo transform with a trainable \ac{vo} network. 
DDVO~\cite{Wang2018} further refined the predicted motion with a differentiable DSO module~\cite{Engel2018}.

Purely monocular approaches are highly sensitive to dynamic objects, which cause incorrect correspondences.
Many works have tried to minimize this impact by introducing predictive masking~\cite{Zhou2017}, uncertainty estimation~\cite{Klodt2018,Yang2020a,Poggi2020}, optical flow~\cite{Yin2018,Ranjan2019,Luo2020} and motion masks~\cite{Gordon2019,Casser2019,Dai2020}.
Monodepth2~\cite{Godard2019} proposed the minimum reconstruction loss and static automasking, encouraging the loss to optimize unoccluded pixels and preventing holes in the depth. 

Other methods focused on the robustness of the photometric loss.
This was achieved through the use of pretrained~\cite{Zhan2018} or learnt~\cite{Spencer2020,Shu2020} feature descriptors and semantic constraints~\cite{Chen2019,Guizilini2020b,Jung2021}.
Mahjourian~\etal~\cite{Mahjourian2018} and Bian~\etal~\cite{Bian2019} complemented the photometric loss with geometric constraints.
ManyDepth~\cite{Watson2019} additionally incorporated the previous frame's prediction into a cost volume.

Complementary to these developments, other works proposed changes to the network architecture, including both the encoder~\cite{Guizilini2020,Bello2021,Zhou2021}, and decoder~\cite{Pillai2019,Guizilini2020,Yan2021,Zhou2021,Lyu2021,Zhao2022}.
Akin to supervised \ac{mde} developments~\cite{Bhat2021,Bhat2022}, Johnston~\etal~\cite{Johnston2020} and Bello~\etal~\cite{Bello2020,Bello2021} obtained improvements by representing depth as a discrete volume.

Finally, several works have complemented self-supervision with proxy depth regression. 
These are typically obtained from \acs{slam}~\cite{Klodt2018,Rui2018}, synthetic data~\cite{Luo2018} or hand-crafted disparity estimation~\cite{Tosi2019,Watson2019}. 
In particular, DepthHints~\cite{Watson2019} improved the proxy depth robustness by generating estimates with multiple hyperparameters. 

The works described here train exclusively on automotive data, such as Kitti~\cite{Geiger2013}, CityScapes~\cite{Cordts2016} or DDAD~\cite{Guizilini2020}.  
Recent benchmark studies~\cite{Spencer2022,Spencer2023} have shown that this lack of variety limits generalization to out-of-distribution domains, such as forests, natural or indoor scenes. 
We propose to greatly increase the diversity and scale of the training data by leveraging unlabelled videos from YouTube, without requiring manual annotation or expensive pre-processing.

%-------------------------------------------------------------------------

% =====================================
%
\begin{table}[t]
\scriptsize
\addtolength{\tabcolsep}{-0.6em}
\renewcommand{\arraystretch}{1.2}
\centering
\input{Tables/datasets}
\label{tbl:data:overview}
\end{table}

% =====================================

% =====================================
%
\begin{figure}[htbp]
\centering
\input{Figures/*}
\label{fig:[}
\end{figure}
t]{slowtv}{data:slowtv}
% =====================================

\section{SlowTV Dataset} \label{sec:data}
\acl{stv} is a style of TV programming featuring uninterrupted shots of long-duration events. 
Our dataset consists of  40 curated videos ranging from 1--8 hours and a total of 135 hours. 

We focus on three categories: hiking, driving and scuba diving.
Hiking videos target natural settings, including forests, mountains or fields, which are non-existent in current datasets. 
These videos were collected in a diverse set of locations and conditions. 
This includes the USA, Canada, the Balkans, Eastern Europe, Indonesia and Hawaii, and conditions such as rain, snow, autumn and summer. 

Existing automotive datasets tend to focus on urban driving in densely populated cities~\cite{Geiger2013,Cordts2016,Huang2019,Chang2019,Guizilini2020,Yu2020,Ceasar2020}. 
Our \acl{stv} dataset features complementary data in the form of long drives in scenic routes, such as mountain and natural trails.
Finally, underwater is an otherwise unused domain, which increases the diversity of the training data and prevents overfitting to purely urban scenes.
\fig{data:slowtv} shows the variability of the proposed dataset, with additional examples and details in \supp{data}.

Videos were downloaded at HD resolution (\shape{720}{1280}{}{}) and extracted at 10 FPS to reduce storage, while still providing smooth motion and large overlap between adjacent frames. 
To make the dataset size tractable and reduce self-similarity, only 100 consecutive frames out of every 250 were retained. 
Despite this, the final training dataset consists of a total of 1.7M images, composed of 1.1M natural, 400k driving and 180k underwater.
\tbl{data:overview} compares existing datasets with those used in this publication.

Since our dataset targets self-supervised methods, the only annotations required are the camera intrinsic parameters. 
We apply COLMAP~\cite{Schoenberger2016} to a sub-sequence to estimate the intrinsics for each video. 
However, as discussed in \sct{meth:cam}, it is possible to let the network jointly optimize camera parameters alongside depth and motion.
This improves performance and results in a truly self-supervised perception and navigation framework, requiring only monocular video to learn how to reconstruct.

%-------------------------------------------------------------------------
\section{Methodology} \label{sec:meth}
\ac{mde} is an alternative to traditional depth estimation techniques, such as stereo matching and cost volumes. 
Rather than relying on multi-view images, these depth networks take only a single image as input.
From this image, a disparity or inverse depth map is estimated as 
% =====================================
$
    \func{\acs{Disp-t}}{\acs{Net-Depth}}{\acs{Img-t}},    
$
% =====================================
where \acs{Net-Depth} represents a trainable \acs{dnn}, \acs{Img-t} is the target image at time-step~\acs{time} and \acs{Disp-t} the predicted sigmoid disparity.

As \acl{stv} contains only monocular videos, we adopt a fully monocular pipeline~\cite{Zhou2017}, whereby our framework also estimates the relative pose~\acs{Pose-tk} between the target~\acs{Img-t} and support frames~\acs{Img-tk}, where 
% =====================================
$ 
    \acs{offset} = \pm1 
$ 
% ===================================== 
is the offset between adjacent frames. 
This is represented as 
% =====================================
$
    \func{\acs{Pose-tk}}{\acs{Net-Pose}}{\acs{Img-t} \concat \acs{Img-tk} },    
$
% =====================================
where \concat is channel-wise concatenation.
Pose is predicted as a translation and axis-angle rotation. 
\looseness=-1

% ..............................................................................
\subsection{Losses} \label{sec:meth:losses}
The correspondences required to warp the support frames and compute the photometric loss are given by backprojecting the depth and re-projecting onto each support frame. 
This process is summarized as 
% =====================================
\begin{equation} \label{eq:reprojection}
    \acs{pix-synth-tk}
    =
    \acs{Cam}
    \acs{Pose-tk}
    \easyfunc{\acs{Depth-t}}{\acs{pix-t}}
    \acs{Cam}^{\inv} 
    \acs{pix-t}
    ,
\end{equation}
% =====================================
where \acs{Cam} are the camera intrinsic parameters, \acs{Depth-t} is the inverted and scaled disparity prediction~\acs{Disp-t}, \acs{pix-t} are the \ndim{2} pixel coordinates in the target frame and \acs{pix-synth-tk} are the reprojected coordinates in the support frame.
We omit the transformation to homogeneous coordinates for simplicity.

The warped support frames are then given by
% =====================================
$
    \acs{Img-synth-tk}
    =
    \acs{Img-tk} \mybil{ \acs{pix-synth-tk} }
    ,
$
% =====================================
where $\mybil{\cdot}$ represents differentiable bilinear interpolation~\cite{Jaderberg2015}.
These warped frames are used to compute the photometric loss \wrt the original target frame.
As is common, we use the weighted combination of SSIM+\pnorm{1}~\cite{Godard2017}, given by
% =====================================
\begin{equation}
    \funcdef
        { \acs{Loss-photo} }
        { \acs{Img}, \acs{Img-synth} }
        { 
            \acs{weight-ssim}
            \frac{ 1 \minus \easyfunc{ \acs{Loss-ssim} }{ \acs{Img}, \acs{Img-synth} } }{2}
            + 
            \mypar{ 1 \minus \acs{weight-ssim} }
            \easyfunc{ \acs{Loss-l1} }{ \acs{Img}, \acs{Img-synth} }
        } 
    ,
\end{equation}
% =====================================
where $\acs{weight-ssim} = 0.85$ is the loss balancing weight.

While Mannequin Challenge consists almost exclusively of static scenes, Kitti and \acl{stv} contain dynamic objects, such as vehicles, hikers, and wild marine life. 
Rather than introducing motion masks~\cite{Gordon2019,Casser2019,Dai2020}, commonly requiring semantic segmentation, we opt for the minimum reconstruction loss~\cite{Godard2019}.
This loss reduces the impact of occluded pixels by optimizing only the pixels with the smallest loss across all support frames and is computed as 
% =====================================
\begin{equation}
    \acs{Loss-rec}
    =
    \asum_{\myac{pix}} 
    \mymin_{\myac{offset}} 
    \easyfunc
        { \acs{Loss-photo} }
        { \acs{Img-t}, \acs{Img-synth-tk} }
    ,
\end{equation}
where $\asum$ indicates averaging over a set.
% =====================================

Finally, automasking~\cite{Godard2019} helps remove holes of infinite depth caused by static frames and objects moving at similar speeds to the camera.
Automasking simply discards pixels where the photometric loss for the \emph{unwarped} target frame is lower than the loss for the synthesized view, given by 
% =====================================
\begin{equation}
    \acs{Mask-static}
    =
    \myivr{ 
        \mymin_{\myac{offset}} 
        \easyfunc{ \acs{Loss-photo} }{ \acs{Img-t}, \acs{Img-synth-tk} }
        <
        \mymin_{\myac{offset}} 
        \easyfunc{ \acs{Loss-photo} }{ \acs{Img-t}, \acs{Img-tk} }
    }
    ,
\end{equation}
% =====================================
where $\myivr{\cdot}$ represents the Iverson brackets.
Additional results showing the effectiveness of the minimum reconstruction loss and automasking can be found in \supp{motion}.

This reconstruction loss is complemented by the common edge-aware smoothness regularization~\cite{Godard2017}.
These networks and losses constitute the core baseline required to train the desired zero-shot depth estimation models. 
To improve existing performance and generalization, we incorporate several new components into the pipeline. 

% ..............................................................................
\subsection{Learning Camera Intrinsics} \label{sec:meth:cam}
As discussed in \sct{data}, we use COLMAP to estimate camera intrinsics for each dataset video.
Whilst this is significantly less computationally demanding than obtaining full reconstructions, it introduces additional pre-processing requirements. 
Eliminating this step would simplify dataset collection and allow for even easier scale-up. 

We take inspiration from~\cite{Gordon2019,Chen2019b} and predict camera intrinsics using the pose network~\acs{Net-Pose}.
This is achieved by adding two decoder branches with the same architecture used to predict pose. 
The modified network is defined as 
% =====================================
$
    \func
        {\acs{Pose-tk}, \acs{fxy}, \acs{cxy}}
        {\acs{Net-Pose}}
        {\acs{Img-t} \concat \acs{Img-tk} },    
$
% =====================================
where \acs{fxy} and \acs{cxy} are the focal lengths and principal point. 

Both quantities are predicted as normalized and scaled by the image shape prior to combining them into \acs{Cam}.
The focal length decoder uses a softplus activation to guarantee a positive output.
The principal point instead uses a sigmoid, under the assumption that it will lie within the image. 
All parameters---depth, pose and intrinsics---are optimized simultaneously, as they all establish the correspondences across support frames, given by \eq{reprojection}.

% ..............................................................................

% =====================================
%
\begin{figure}[t]
\centering
\input{Figures/distort}
\label{fig:meth:distort}
\end{figure}

% =====================================

\subsection{Aspect Ratio Augmentation} \label{sec:meth:aug}
The depth network is commonly a fully convolutional network that can process images of any size. 
In practice, these networks can overfit to the training size, resulting in poor out-of-dataset performance. 
\fig{meth:distort} shows this effect, where resizing to the training resolution improves results, despite introducing stretching or squashing distortions. 

Since both \acl{stv} and \acl{mc} were sourced from YouTube, they feature the common widescreen aspect ratio (16:9). 
However, the objective is to train a model that can be easily applied to real-world settings in a zero-shot fashion. 
To this end, we propose an \ac{araug} that randomizes the image shape during training, increasing the data diversity.

\ac{araug} has two components: centre cropping and resizing. 
The cropping stage uniformly samples from a set of predefined aspect ratios. 
A random crop is generated using this aspect ratio, covering 50-100\% of the original height or width.
By definition, the sampled crop will be smaller than the original image and of different shape. 
The crop is therefore resized to match the number of pixels in the original image. 
\supp{aug} details the full set of aspect ratios used and shows training images obtained using this procedure.

\ac{araug} has the effect of drastically increasing the distribution of image shapes, aspect ratios and object scales seen by the network during training.
As shown in \sct{res:abl}, this greatly increases performance, especially when evaluating on datasets with different image sizes.

%-------------------------------------------------------------------------
\section{Results} \label{sec:res}
We evaluate the proposed models in a variety of settings and datasets, including in-distribution and zero-shot. 
Since the trained model is purely monocular, the predicted depth is in arbitrary units. 
Instead of using traditional median alignment~\cite{Zhou2017,Godard2019}, we follow \acl{midas}~\cite{Ranftl2020} and estimate scale and shift alignment parameters based on a least-squares criterion. 
We apply the same strategy to every baseline.
Results using median alignment are shown in \supp{res}.
Note that datasets with \ac{sfm} ground-truth (\eg \acl{mc}) are also scaleless and would require this step even for techniques that predict metric depth. 

% ..............................................................................
\subsection{Implementation Details} \label{sec:res:details}
The proposed models are implemented in PyTorch~\cite{Paszke2019} using the baselines from the Monodepth Benchmark~\cite{Spencer2022}.
The depth network uses a pretrained ConvNeXt{\nbd}B backbone~\cite{Liu2022,Wightman2019} and a DispNet decoder~\cite{Mayer2016,Godard2017}.
The pose network instead uses ConvNeXt{\nbd}T for efficiency. 
Each model variant is trained with three random seeds and we report average performance.
This improves the reliability of the results and reduces the impact of non-determinism.

The final models were trained on a combination of \acl{stv} (1.7M), \acl{mc} (115k) and \acl{keb} (71k). 
To make the duration of each epoch tractable and balance the contribution of each dataset, we fix the number of images per epoch to 30k, 15k and 15k, respectively. 
The subset sampled from each dataset varies with each epoch to ensure a high data diversity.  

The models were trained for 60 epochs using AdamW~\cite{Loshchilov2017} with weight decay $10^{-3}$ and a base learning rate of $10^{-4}$, decreased by a factor of 10 for the final 20 epochs.
Empirically, we found that linearly warming up the learning rate for the first few epochs stabilized learning and prevented model collapse. 
We use a batch size of 4 and train the models on a single NVIDIA GeForce RTX 3090.

\acl{stv} and \acl{mc} use a base image size of \shape{384}{640}{}{}, while Kitti uses \shape{192}{640}{}{}.
As is common, we apply horizontal flipping and colour jittering augmentations with 50\% probability. 
\ac{araug} is applied with 70\% probability, sampling from 16 predefined aspect ratios. 
The full set of aspect ratios can be found in \supp{aug}. 

Since existing models are trained exclusively on automotive data, most of the motion occurs in a straight-line and forward-facing direction. 
It is therefore common practice to force the network to always make a forward-motion prediction by reversing the target and support frame if required. 
Handheld videos, while still primarily featuring forward motion, also exhibit more complex motion patterns.
As such, removing the forward motion constraint results in a more flexible model that improves performance. 

Similarly, existing models are trained with a fixed set of support frames---usually previous and next.
Since \acl{stv} and \acl{mc} are mostly composed of handheld videos, the change from frame-to-frame is greatly reduced. 
We make the model more robust to different motion scales and appearance changes by randomizing the separation between target and support frames. 
In general, we sample such that handheld videos use a wider time-gap between frames, while automotive has a small time-gap to ensure there is significant overlap between frames. 
As shown later, this leads to further improvements and greater flexibility. 

% ====================================
%
\begin{table}[b]
\scriptsize
\addtolength{\tabcolsep}{-0.6em}
\renewcommand{\arraystretch}{1.2}
\centering
\input{Tables/resources}
\label{tbl:res:resources}
\end{table}

% ====================================

% ..............................................................................
\subsection{Baselines} \label{sec:res:baselines}
We use the \acs{ssl} baselines from~\cite{Spencer2022}, trained on \acl{kez} with a ConvNeXt{\nbd}B backbone. 
We minimize architecture changes and training settings \wrt the baselines to ensure models are comparable and improvements are solely due to the contributions from this paper.

We also report results for recent \ac{sota} supervised \ac{mde} approaches, namely \acl{midas}~\cite{Ranftl2020}, DPT~\cite{Ranftl2021} and \acl{crf}~\cite{Weihao2022}. 
\acl{midas} and DPT were trained on a large collection of supervised datasets that do not overlap with our testing datasets (unless otherwise indicated).
As such, these models are also evaluated in a zero-shot fashion. 
We use the pre-trained models and pre-processing provided by the PyTorch Hub.
\acl{crf} provides separate indoor/outdoor models, trained on Kitti and \acl{nyud} respectively.
We evaluate the corresponding model in a zero-shot manner depending on the dataset category. 
Despite predicting metric depth, we apply scale and shift alignment to ensure results are comparable.

% ..............................................................................
\subsection{Evaluation Metrics} \label{sec:res:metrics}

We report the following metrics per dataset:

\heading{\acs{rel}} 
Absolute relative error (\%) between target \acs{target} and prediction \acs{pred} as 
$
    \text{\acs{rel}} = \asum \mymag{\acs{target} \minus \acs{pred}} / \acs{target}.
$

\heading{Delta}
Prediction threshold accuracy (\%) as \\
$
    \acs{d25} = \asum \mypar{ \max \mypar{ \acs{pred}/\acs{target},\ \acs{target}/\acs{pred} } < 1.25 }.
$

\heading{\acs{fscore}}
Pointcloud reconstruction F-Score~\cite{Ornek2022} (\%) as 
$
    \text{\acs{fscore}} = \mypar{2 P R}/\mypar{P + R},
$
where $P$ and $R$ are the Precision and Accuracy of the \ndim{3} reconstruction with a correctness threshold of 10cm.

\vspace{0.2cm}
We additionally compute multi-task metrics to summarize the performance across all datasets: 

\heading{\acs{rank}}
Average ordinal ranking order across all metrics as 
$
    \text{\acs{rank}} = \asum_{m} r_{m},
$
where $m$ represents each available metric and $r$ is the ordinal rank. 

\heading{Improvement}
Average relative performance increase (\%) across all metrics as 
$
    \acs{relimp} = \asum_{m} (\minus 1)^{l_m} (M_m - M^0_m) / M^0_m,
$
where $l_m = 1$ if lower is better, $M_m$ is the performance for a given metric and $M^0_m$ is the baseline's performance.

% ..............................................................................

% =====================================
%
\begin{table}[b]
\scriptsize
\addtolength{\tabcolsep}{-0.6em}
\renewcommand{\arraystretch}{1.2}
\centering
\input{Tables/res_abl}

\label{tbl:res:abl}
\end{table}

% =====================================

\subsection{Ablation} \label{sec:res:abl}
We perform a ``leave-one-out'' ablation study, whereby a single component is removed per-experiment from the full model. 
This helps to understand the impact of each proposed contribution. 
We report this ablation on \acl{kez}, \acl{mc} and \acl{syns}.

As shown in \tbl{res:abl}, the full model with all contributions performs best.
\emph{Fwd \acs{Pose}} represents a network forced to always predict forward-motion. 
\emph{$\acs{offset} = \pm1$} uses fixed support frames, instead of the randomization in \sct{res:details}.
\emph{Fixed \acs{Cam}} removes the learnt intrinsics from \sct{meth:cam}, while \emph{No \acs{araug}} removes the aspect ratio augmentation. 
It is worth noting that none of these contributions increase the number of depth network parameters.
Learning the intrinsics results in a negligible increase in the pose network, which is not required for inference. 
Despite this, each contribution significantly improves accuracy and generalization.  

% ..............................................................................

% =====================================
%
\begin{table}[htbp]
\scriptsize
\addtolength{\tabcolsep}{-0.6em}
\renewcommand{\arraystretch}{1.2}
\centering
\input{Tables/*}
\label{tbl:[}
\end{table}
t]{res}{res:all}
% =====================================

% =====================================
%
\begin{figure}[htbp]
\centering
\input{Figures/*}
\label{fig:[}
\end{figure}
t]{preds}{res:viz:zero}
% =====================================

\subsection{In-distribution} \label{sec:res:base}
We compare our best approach---\ac{kbr}---against existing \ac{sota} on the two training datasets with ground-truth: Kitti and \acl{mc}. 
This represents the most common evaluation, where the test data is sampled from the same distribution as the training data. 

As shown in \tbl{res:all} (\emph{In-Distribution}), all variants of the proposed models outperform the improved \ac{ssl} baselines from~\cite{Spencer2022}.
Even more surprising, our models also outperform most \emph{supervised} baselines on Kitti, despite \acl{dpt-beit} being trained on it.
\acl{crf} is the only supervised model to outperform ours by a slight margin. 
This may be  due to the additional automotive data from \acl{stv}, which increases the variety and improves generalization. 
Finally, our model outperforms even the supervised \ac{sota} on \acl{mc} \acl{fscore}.

% ..............................................................................
\subsection{Zero-shot Generalization} \label{sec:res:zero}
The core of our evaluation takes place in a \emph{zero-shot} setting, \ie models are not fine-tuned.
This demonstrates the capability of our model to generalize to previously unseen environments.
While several existing \acl{ssmde} approaches provide zero-shot evaluations, this is usually limited to CityScapes~\cite{Cordts2016} and Make3D~\cite{Saxena2009}.
These datasets provide low-quality ground-truths and focus exclusively on urban environments similar to Kitti.
We instead opt for a collection of challenging datasets, constituting a mixture of urban, natural, synthetic and indoor scenes.

\heading{Outdoor}
These results can be found in \tbl{res:all} (\emph{Outdoor}), where all evaluated models are zero-shot.
Once again, our models outperform the \ac{ssl} baselines in every metric, across all datasets. 
\acl{crf} is capable of generalizing to other automotive datasets and provides good performance on \acl{ddad}.
However, our model adapts better to complex synthetic (\acl{sintel}) and natural (\acl{syns}) scenes.
Despite being fully self-supervised and requiring no depth annotations during training, our model outperforms \acl{midas} and \acl{dpt-vit}.
DPT leverages expensive transformer-based backbones and additional datasets to improve performance.

\heading{Indoor}
\tbl{res:all} (\emph{Indoor}) shows results for all indoor datasets. 
Note that \acl{crf} was trained exclusively on \acl{nyud}, while \acl{dpt-beit} used it as part of its training collection.
As such, this subset of results is \emph{not} zero-shot.
As with the outdoor evaluations, our model provides significant improvements over all existing \ac{ssl} approaches. 
This is due to the focus on Kitti and the lack of indoor training data, highlighting the need for more varied training sources.
However, the supervised models still provide improvements over our method, likely due to the additional indoor datasets used for training.
Once again, we emphasize that our model is \emph{fully self-supervised}. 
Despite this, we close the performance gap on complex supervised models.

\heading{Visualizations}
We visualize the network predictions in \fig{res:viz:zero}. 
As seen, the proposed model clearly outperforms the best \ac{ssl} baseline.
This is most noticeable in indoor settings, where the baseline treats human faces as background. 
In many cases, our self-supervised model provides similar or better depth maps than the supervised baselines. 
Once again, these rely on ground-truth annotations and expensive transformer-based backbones.
Meanwhile, our model simply requires curated collections of freely-available monocular YouTube videos, without even camera intrinsics.

\heading{Failure Cases}
Our approach does not explicitly use explicit motion masks to handle dynamic objects. 
Instead, we rely only on the minimum reconstruction loss and automasking~\cite{Godard2019}.
Whilst this improves the robustness, it can be seen how dynamic objects such as cars can cause incorrect predictions (\eg Kitti or \acs{ddad}). 
This represents one of the most important avenues for future research. 
Further discussions regarding these failure cases and additional visualizations can be found in \supp{fail}.

\heading{MDEC-2}
The Monocular Depth Estimation Challenge~\cite{Spencer2023,Spencer2023b} tested zero-shot generalization on \acl{syns}. 
We compare our model to all submissions from the latest edition (CVPR2023).
As seen in \fig{res:mdec}, our method (\textit{jspenmar2}) achieves the highest F-Score reconstruction and is top-3 in all metrics except AbsRel and Edge-Accuracy. 
Once again, this illustrates the benefits of \acl{stv}, which contains large quantities of natural data not present in other datasets.

% =====================================
%
\begin{figure}[t]
\centering
\input{Figures/mdec}
\label{fig:res:mdec}
\end{figure}

% =====================================

% ..............................................................................
\subsection{Map-Free Relocalization} \label{sec:res:reloc}
Map-free relocalization is the task of localizing a target image using a single reference image.
This is contrary to traditional pipelines, which require large image collections to first build a scene-specific map, such as \ac{sfm} or training a \acs{cnn}. 
Recent work~\cite{Toft2020,Arnold2022} has shown the benefit of incorporating metric \ac{mde} into feature matching pipelines to resolve the ambiguous scale of the predicted pose.

We evaluate all depth models on the MapFreeReloc benchmark~\cite{Arnold2022} validation split, serving as an example real-world task.
The feature-matching baseline~\cite{Arnold2022} consists of LoFTR~\cite{Sun2021} correspondences, a PnP solver and DPT~\cite{Ranftl2021} fine-tuned on either Kitti or \acl{nyud}.
Since this benchmark requires metric depth but does not provide ground-truth, we align all models to the baseline fine-tuned DPT predictions using least-squares. 
We report the metrics provided by the benchmark authors. 
This includes translation (meters), rotation (deg) and reprojection (px) errors.
Pose Precision/AUC were computed with an error threshold of 25 cm \& 5$^\circ$, while Reprojection uses a threshold of 90px.

As shown in \tbl{res:reloc}, our method has the best performance across all \acl{ssmde} approaches by a large margin.
Our performance is on par with the supervised \ac{sota}, without  requiring ground-truth supervision.
This further demonstrates the benefits of the proposed \acl{stv} dataset and its applicability to real-world scenarios. 
Interestingly, we find that the original DPT models perform better than their fine-tuned counterparts, despite using these as the metric scale reference. 
This suggests that the fine-tuning procedure of~\cite{Arnold2022} may provide metric scale at the cost of generality.  
However, this highlights the need for models that predict accurate metric depth, rather than only relative depth.

% =====================================
%
\begin{table}[t]
\scriptsize
\addtolength{\tabcolsep}{-0.6em}
\renewcommand{\arraystretch}{1.2}
\centering
\input{Tables/res_reloc}
\label{tbl:res:reloc}
\end{table}

% =====================================

%-------------------------------------------------------------------------
\section{Conclusion} \label{sec:conc}
This paper has presented the first approach to \acl{ssmde} capable of generalizing across many datasets, including a wide range of indoor and outdoor environments.
We demonstrated that our models significantly outperform existing self-supervised models, even in the automotive domain where they are currently trained.
By leveraging the large quantity and variety of data in the new \acl{stv} dataset, we are able to close the gap between supervised and self-supervised performance.
Additional components, such as the novel \ac{araug}, randomized support frames and more flexible pose estimation, further improve the performance and zero-shot generalization of the proposed models.

Future work should explore alternative sources of data to incorporate even more scene variety. 
In particular, additional indoor data may significantly reduce the remaining gap between self-supervised and supervised approaches.  
Another key direction is improving the accuracy in dynamic scenes. 
A promising approach would be using optical flow to refine the estimated correspondences. 
This could be incorporated in a self-supervised manner, without requiring semantic segmentation or motion masks. 
However, it introduces additional costs due to the increased computational requirements from the new network. 

Developing models capable of predicting metric depth would further increase their applicability to real-world applications.
Finally, as the diversity of training environments increases, it will become crucial to further diversify the benchmarks used to evaluate these models.

%-------------------------------------------------------------------------
\subsection*{Acknowledgements}
This work was partially funded by the EPSRC under grant agreements EP/S016317/1 \& EP/S035761/1.

%-------------------------------------------------------------------------
\include{appendices}

%-------------------------------------------------------------------------
{\small
\bibliographystyle{ieee_fullname}
\bibliography{8981}
}

\end{document}

%% file: Preamble/preamble.tex
\usepackage{acronym}
\usepackage{adjustbox}
\usepackage{booktabs}
\usepackage[font=small,labelfont=bf]{caption}
\usepackage{colortbl}
\usepackage{dashrule}
\usepackage{fmtcount}
\usepackage{forloop}
\usepackage{stackengine}
\usepackage{subfig}
\usepackage{suffix}
\usepackage{svg}
\usepackage{rotating}
\usepackage{pifont}    % ding
\usepackage{stmaryrd}  % iverson brackets
\usepackage{tabularx}
\usepackage[textsize=scriptsize]{todonotes}
\usepackage{xfrac}
\usepackage{xr}
\usepackage{xspace}

\presetkeys{todonotes}{inline}{}  % Inline by default

\makeatletter
\newcommand*{\addFileDependency}[1]{% argument=file name and extension
  \typeout{(#1)}
  \@addtofilelist{#1}
  \IfFileExists{#1}{}{\typeout{No file #1.}}
}
\makeatother

\definecolor{ForestGreen}{HTML}{228b22}  % xcolor tends to have option clash

\DeclareRobustCommand{\nbd}{\nobreakdash-} % Usage: "multi{\nbd}scale"

\newcommand{\cmark}{\textcolor{ForestGreen}{\ding{51}}}
\newcommand{\xmark}{\textcolor{red}{\ding{55}}}
\newcommand{\mydag}{\textcolor{red}{\textsuperscript{\bf\textdagger}}}

\newcommand{\down}{\textcolor{black}{\ensuremath{\boldsymbol{\downarrow}}}}
\newcommand{\up}{\textcolor{black}{\ensuremath{\boldsymbol{\uparrow}}}}

\newcommand{\ndim}[1]{#1{\nbd}D}
\newcommand{\pnorm}[1]{L\ensuremath{_#1}}

\newcommand{\best}[1]{\textcolor{ForestGreen}{\textbf{#1}}}
\newcommand{\nbest}[1]{\textcolor{blue}{\underline{#1}}}
\newcommand{\zs}{\cellcolor{red!25}}

\newcommand{\heading}[1]{\noindent \textbf{\small #1.}}
\newcommand{\mycaption}[2]{\caption[#1]{\textbf{#1.} #2}}

\newcommand{\addfig}[3][htbp]{%
\begin{figure}[#1]
\centering
\input{Figures/#2}
\label{fig:#3}
\end{figure}
}

\WithSuffix\newcommand\addfig*[3][htbp]{%
\begin{figure*}[#1]
\centering
\input{Figures/#2}
\label{fig:#3}
\end{figure*}
}

\newcommand{\addtbl}[3][htbp]{%
\begin{table}[#1]
\scriptsize
\addtolength{\tabcolsep}{-0.6em}
\renewcommand{\arraystretch}{1.2}
\centering
\input{Tables/#2}
\label{tbl:#3}
\end{table}
}

\WithSuffix\newcommand\addtbl*[3][htbp]{%
\begin{table*}[#1]
\scriptsize
\addtolength{\tabcolsep}{-0.6em}
\renewcommand{\arraystretch}{1.2}
\centering
\input{Tables/#2}
\label{tbl:#3}
\end{table*}
}

\def\eg{\emph{e.g}.~} 
\def\ie{\emph{i.e}.~}

\def\wrt{w.r.t.~} 

\def\etal{\emph{et al}.\xspace}

\newcommand{\fig}[1]{Figure~\ref{fig:#1}}
\newcommand{\tbl}[1]{Table~\ref{tbl:#1}}
\newcommand{\sct}[1]{Section~\ref{sec:#1}}
\newcommand{\supp}[1]{Appendix~\ref{sec:supp:#1}}
\newcommand{\eq}[1]{(\ref{eq:#1})}

\newcommand{\acromath}[3]{\acrodef{#1}[\(#2\)]{#3}} 
\newcommand{\myac}[1]{\text{\acs{#1}}}  % For inserting acronyms in acronyms or as subscripts!

\DeclareRobustCommand{\asum}{\ensurestackMath{\mathop{\mathchoice%
  {\stackon[-3.8ex]{\displaystyle\sum}{\smash{\rule{.4pt}{4ex}}}}%
  {\stackon[-2.6ex]{\textstyle\sum}{\smash{\rule{.4pt}{2.9ex}}}}%
  {\stackon[-1.9ex]{\scriptstyle\sum}{\smash{\rule{.4pt}{2.2ex}}}}%
  {\stackon[-1.4ex]{\scriptscriptstyle\sum}{\smash{\rule{.4pt}{1.7ex}}}}%
}}}

\DeclareMathOperator*{\mymin}{min}

\newcommand{\minus}{\scalebox{0.5}[0.75]{\ensuremath{-}}}
\newcommand{\inv}{\minus 1}

\newcommand{\shape}[4]{\ensuremath{  % a x b x c x d
#1 \times #2
\ifthenelse {\equal{#3}{}} {} {\times #3}
\ifthenelse {\equal{#4}{}} {} {\times #4}
}}
\newcommand{\sca}[1]{\ensuremath{#1}}  						   % Non-bold
\newcommand{\vct}[1]{\ensuremath{\textbf{\MakeLowercase{#1}}}} % Lowercase/Bold
\newcommand{\mat}[1]{\ensuremath{\textbf{\MakeUppercase{#1}}}} % Uppercase/Bold
\newcommand{\weight}{\lambda}

\newcommand{\Loss}{\mathcal{L}}
\newcommand{\Mask}{\ensuremath{\mathbb{M}}}

\newcommand{\concat}{\ensuremath{\oplus}\xspace}

\newcommand{\brackets}[3]{\ensuremath{\left#1 #2 \right#3}}
\newcommand{\mypar}[1]{\brackets{(}{#1}{)}}

\newcommand{\mymag}[1]{\brackets{|}{#1}{|}}  

\newcommand{\mybil}[1]{\brackets{\langle}{#1}{\rangle}}  % < a >
\newcommand{\myivr}[1]{\brackets{\llbracket}{#1}{\rrbracket}}  % [[ a ]]

\newcommand{\easyfunc}[2]{\ensuremath{#1\!\mypar{#2}}}                 % a(b)
\newcommand{\func}[3]{\ensuremath{ #1 = \easyfunc{#2}{#3} }}         % a = b(c)
\newcommand{\funcdef}[3]{\ensuremath{ \easyfunc{#1}{#2} = #3 }}      % a(b) = c

%% file: Preamble/acronyms.tex
\acrodef{ar}     [AR]             {Augmented Reality}
\acrodef{araug}  [AR-Aug]         {aspect ratio augmentation}

\acrodef{cnn}    [CNN]            {Convolutional Neural Network}

\acrodef{d25}    [$\delta_{.25}$] {$\delta < 1.25$}
\acrodef{dnn}    [DNN]            {Deep Neural Network}

\acrodef{fscore} [F]              {F-Score}

\acrodef{imu}    [IMU]            {Inertial Measurement Unit}

\acrodef{kbr}    [KBR]            {Kick Back \& Relax}

\acrodef{lidar}  [LiDAR]          {Light Detection and Ranging}

\acrodef{mae}    [MAE]            {Mean Absolute Error}
\acrodef{mde}    [MDE]            {monocular depth estimation}
\acrodef{ml}     [ML]             {Machine Learning}

\acrodef{rank}    [Rank]            {Mean Rank}
\acrodef{rel}    [Rel]            {Absolute Relative Error}
\acrodef{relimp} [$\Delta$]       {Relative Improvement}

\acrodef{sfm}    [SfM]            {Structure-from-Motion}
\acrodef{slam}   [SLAM]           {Simultaneous Localization and Mapping}
\acrodef{sota}   [SotA]           {State-of-the-Art}
\acrodef{ssl}    [SSL]            {self-supervised learning}

\acrodef{vo}     [VO]             {Visual Odometry}

\acrodef{crf}      [NeWCRFs] {NeWCRFs}

\acrodef{ddad}     [DDAD]    {DDAD}
\acrodef{dii}      [DIODE]   {DIODE Indoors}
\acrodef{dio}      [DIODE]   {DIODE Outdoors}
\acrodef{dpt-beit} [DPT]     {DPT-BEiT}
\acrodef{dpt-vit}  [DPT]     {DPT-ViT}

\acrodef{kb}       [KB]      {Kitti Benchmark}
\acrodef{ke}       [KE]      {Kitti Eigen}
\acrodef{keb}      [Kitti]     {Kitti Eigen{\nbd}Benchmark}
\acrodef{kez}      [KEZ]     {Kitti Eigen{\nbd}Zhou}

\acrodef{mc}       [Mannequin]      {Mannequin Challenge}
\acrodef{midas}    [MiDaS]   {MiDaS}

\acrodef{nyud}     [NYUD]    {NYUD{\nbd}v2}

\acrodef{sintel}   [Sintel]  {Sintel}
\acrodef{ssmde}    [SS-MDE]  {SS{\nbd}MDE}
\acrodef{stv}      [STV]     {SlowTV}
\acrodef{syns}     [SYNS]    {SYNS{\nbd}Patches}

\acrodef{tum}      [TUM]     {TUM{\nbd}RGBD}

%% file: Preamble/symbols.tex
\acromath{Cam} {\mat{K}} {}

\acromath{fxy} {\vct{f}_{xy}} {}
\acromath{cxy} {\vct{c}_{xy}} {}

\acromath{Depth} {\mat{D}} {}
\acromath{Depth-t} {\myac{Depth}_{\myac{time}}} {}
\acromath{Disp} {\hat{\ac{Depth}}} {}
\acromath{Disp-t} {\myac{Disp}_{\myac{time}}} {}
\acromath{Disp-tk} {\myac{Disp}_{\myac{tk}}} {}

\acromath{Img} {\mat{I}} {}
\acromath{Img-synth} {\myac{Img}'} {}
\acromath{Img-synth-tk} {\myac{Img-synth}_{\myac{tk}}} {}
\acromath{Img-t} {\myac{Img}_{\myac{time}}} {}
\acromath{Img-tk} {\myac{Img}_{\myac{tk}}} {}

\acromath{Loss-l1}{\sca{\Loss}_{\textit{1}}}{}
\acromath{Loss-ssim}{\sca{\Loss}_{\textit{ssim}}}{}
\acromath{Loss-photo}{\sca{\Loss}_{\textit{ph}}}{}
\acromath{Loss-rec}{\sca{\Loss}_{\textit{rec}}}{}

\acromath{Mask-static}{\mat{\Mask}}{}

\acromath{Net-Depth} {\sca{\Phi}_{\textit{D}}} {}
\acromath{Net-Pose} {\sca{\Phi}_{\textit{P}}} {}

\acromath{offset} {\sca{k}} {}
\acromath{pix} {\vct{p}} {}
\acromath{pix-t} {\myac{pix}_{\myac{time}}} {}
\acromath{pix-synth} {\myac{pix}'} {}
\acromath{pix-synth-tk} {\myac{pix-synth}_{\myac{tk}}} {}
\acromath{Pose} {\hat{\mat{P}}} {}
\acromath{Pose-tk} {\myac{Pose}_{\myac{tk}}} {}

\acromath{time} {\sca{t}} {}
\acromath{tk}{\myac{time}\!+\!\myac{offset}} {}

\acromath{weight-ssim}{\sca{\weight}}{}

\acromath{target}{\sca{y}} {}
\acromath{pred}{\hat{\myac{target}}} {}

%% file: Figures/intro.tex
% Intro banner teaser.

\captionsetup{type=figure}

\subfloat{\includegraphics[width=0.25\linewidth]{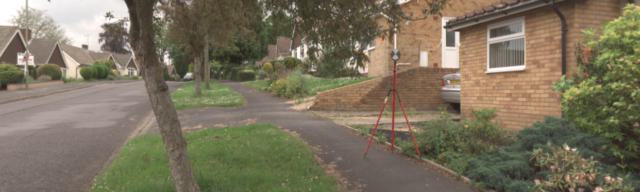}}%
\subfloat{\includegraphics[width=0.25\linewidth]{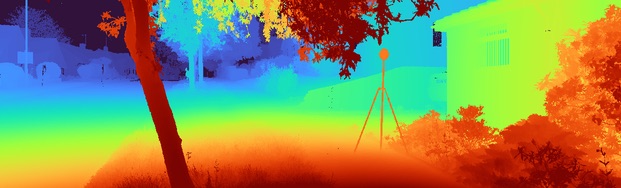}}%
\subfloat{\includegraphics[width=0.25\linewidth]{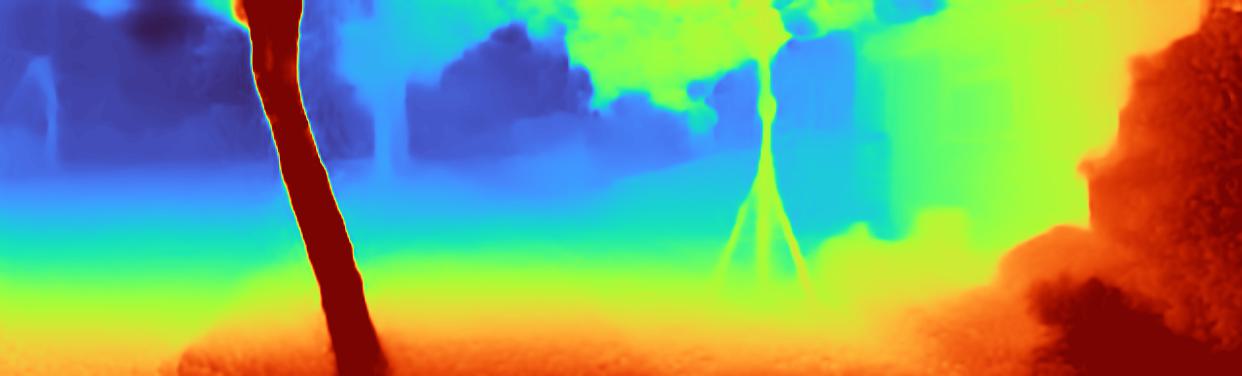}}
\subfloat{\includegraphics[width=0.25\linewidth]{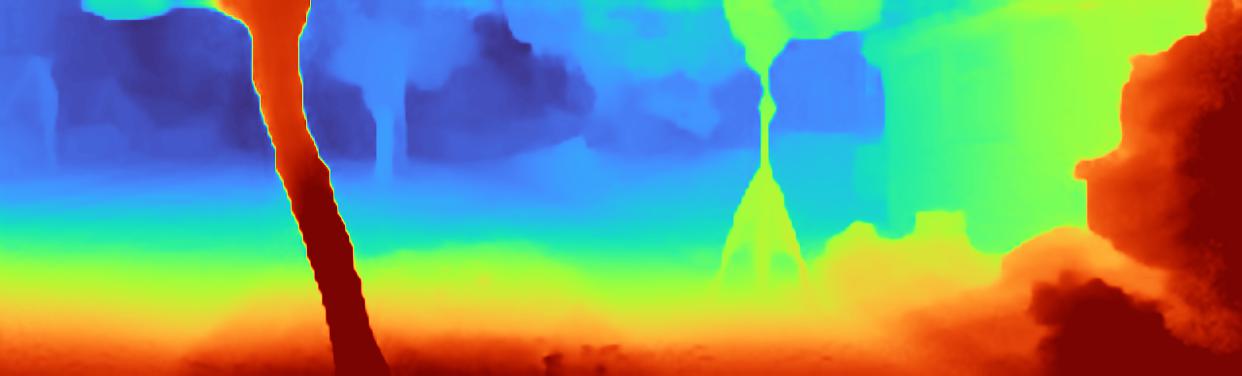}}%
\\[-2.5ex]
\subfloat[\textbf{Image}]{\includegraphics[width=0.25\linewidth]{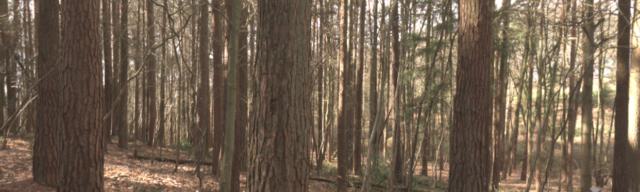}}%
\subfloat[\textbf{Ground-truth}]{\includegraphics[width=0.25\linewidth]{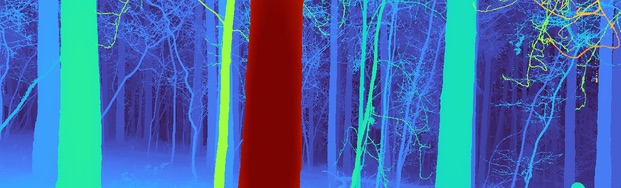}}%
\subfloat[\textbf{Ours}]{\includegraphics[width=0.25\linewidth]{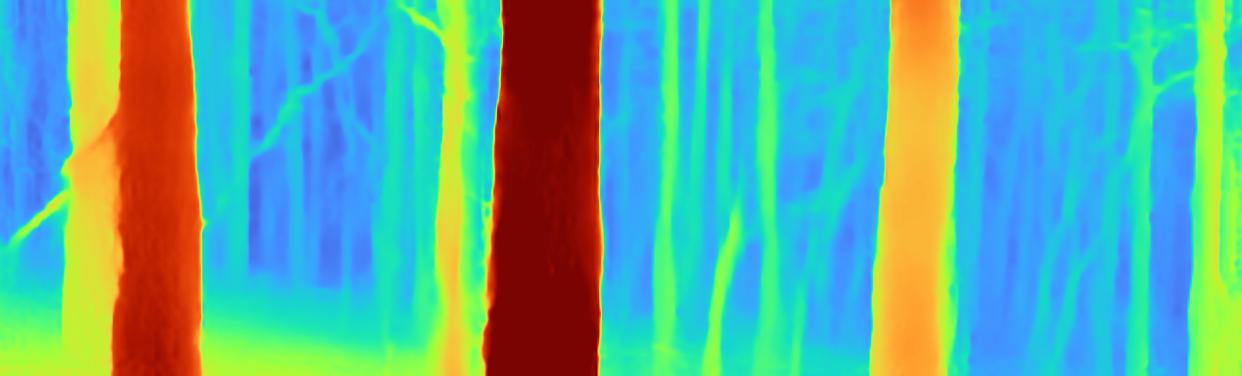}}%
\subfloat[\textbf{Baseline~\cite{Spencer2022}}]{\includegraphics[width=0.25\linewidth]{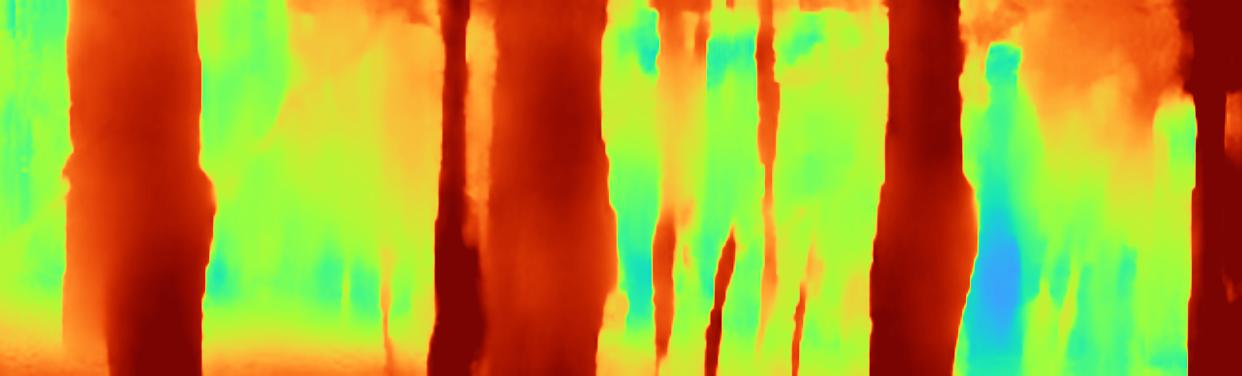}}%

\captionof{figure}{\textbf{Zero-shot Generalization.}
We present the first \acl{ssmde} model capable of generalizing to a wide-range of complex environments.
This is achieved by training on the novel large-scale \acl{stv} dataset.
We outperform other existing self-supervised methods and perform on par with recent supervised \acs{sota}~\cite{Ranftl2020,Ranftl2021,Weihao2022}.
}

%% file: Tables/datasets.tex
% Summary of training & eval datasets.

\mycaption{Datasets Comparison}{%
The top half shows commonly used \acl{ssmde} training datasets.
The proposed \acl{stv} greatly diversifies training environments and scales to much larger quantities.
The bottom half summarizes the testing datasets used in our \emph{zero-shot generalization} evaluation.
}

\begin{tabular}{lccccccccc}
\toprule
                                        &  Urban & Natural &  Scuba & Indoor &       Depth &    Acc & Density &         \#Img \\
\midrule      
Kitti~\cite{Geiger2013,Uhrig2018}\mydag & \cmark &  \xmark & \xmark & \xmark & \acs{lidar} &   High &     Low &           71k \\
\acl{ddad}~\cite{Guizilini2020}         & \cmark &  \xmark & \xmark & \xmark & \acs{lidar} &    Mid &     Low &           76k \\
CityScapes~\cite{Cordts2016}            & \cmark &  \xmark & \xmark & \xmark &      Stereo &    Low &     Mid &           88k \\
Mannequin~\cite{Li2020}\mydag           & \cmark &  \xmark & \xmark & \cmark &    \ac{sfm} &    Mid &     Mid &          115k \\
\textbf{\acl{stv} (Ours)}\mydag         & \cmark &  \cmark & \cmark & \xmark &      \xmark & \xmark &  \xmark & \textbf{1.7M} \\
\midrule                             
Kitti~\cite{Geiger2013,Uhrig2018}       & \cmark &  \xmark & \xmark & \xmark & \acs{lidar} &   High &     Low &           652 \\
\acl{ddad}~\cite{Guizilini2020}         & \cmark &  \xmark & \xmark & \xmark & \acs{lidar} &    Mid &     Low &            1k \\
\acl{sintel}~\cite{Butler2012}          & \xmark &  \cmark & \xmark & \xmark &       Synth &   High &    High &          1064 \\
\acl{syns}~\cite{Adams2016,Spencer2022} & \cmark &  \cmark & \xmark & \cmark & \acs{lidar} &   High &    High &           775 \\
\acs{dii}~\cite{Vasiljevic2019}         & \cmark &  \xmark & \xmark & \cmark & \acs{lidar} &   High &    High &           771 \\
Mannequin~\cite{Li2020}                 & \cmark &  \xmark & \xmark & \cmark &    \ac{sfm} &    Mid &     Mid &            1k \\
\acl{nyud}~\cite{Silberman2012}         & \xmark &  \xmark & \xmark & \cmark &      Kinect &    Mid &    High &           654 \\
\acl{tum}~\cite{Sturm2012}              & \xmark &  \xmark & \xmark & \cmark &      Kinect &    Mid &    High &          2.5k \\
\bottomrule
\end{tabular}

\emph{%
\mydag Datasets used to train our networks.
}

\vspace*{-0.3cm}

%% file: Figures/distort.tex
% Example of distorting image to training size.

\subfloat[Image]{\includegraphics[width=0.49\linewidth]{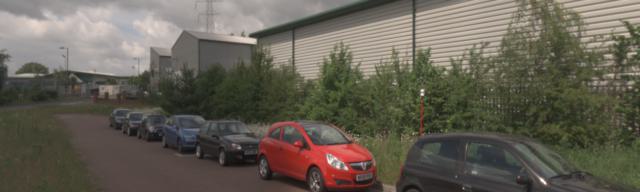}} \hfill
\subfloat[Ground-truth]{\includegraphics[width=0.49\linewidth]{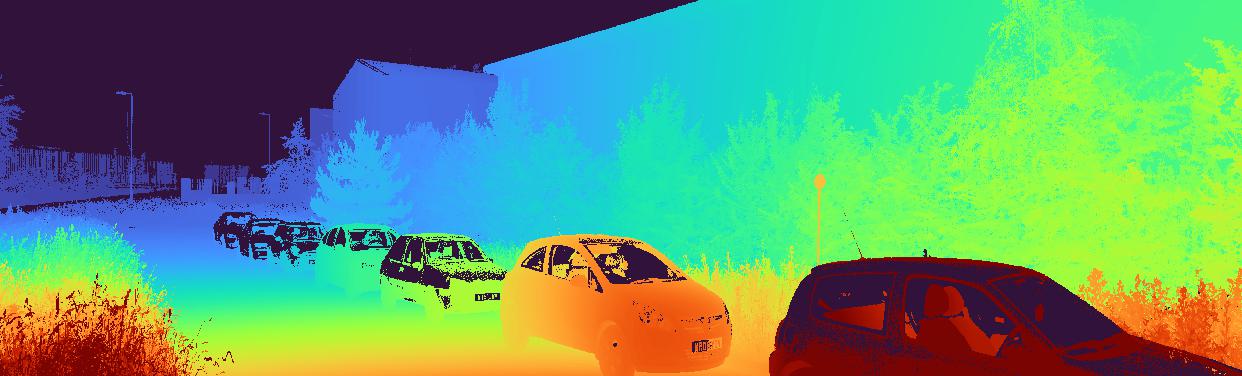}} 
\\[-2ex]
\subfloat[Base ($\delta_{.25} = 61.82\%$)]{\includegraphics[width=0.49\linewidth]{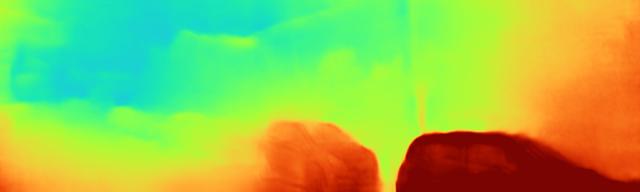}}  \hfill
\subfloat[Distorted ($\delta_{.25} = \best{71.12\%}$)]{\includegraphics[width=0.49\linewidth]{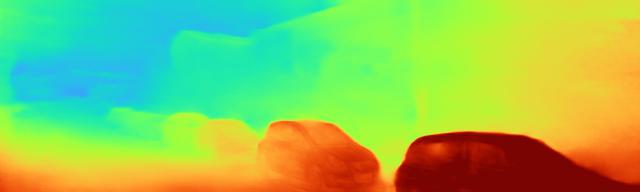}} 

\vspace*{-0.2cm}

\mycaption{Generalizing to Image Shapes}{%
The same model, at different resolutions, can produce significantly different predictions. 
Distorting the image (and resizing the prediction) can improve performance, despite introducing artefacts. 
Note the improved boundary sharpness in (d).
}

\vspace*{-0.5cm}

%% file: Tables/resources.tex
% Model params & FPS.

\mycaption{Model Complexity}{%
Supervised \acs{sota} approaches make use of computationally expensive transformer backbones. 
Despite being of equivalent complexity to the \ac{ssl} baselines~\cite{Spencer2022}, our model closes the gap to supervised performance.
}

\begin{tabular}{@{}llllll@{}}
\toprule
                          & Backbone                     & MParam\down    & FPS\up        \\
\midrule 
\textbf{\acs{kbr} (Ours)} & ConvNeXt-B~\cite{Liu2022}    & \best{92.65}   & \best{61.50}  \\
\midrule
MiDaS~\cite{Ranftl2020}   & ResNeXt-101~\cite{Xie2017}   & \nbest{105.36} & \nbest{51.38} \\
DPT~\cite{Ranftl2021}     & ViT-L~\cite{Dosovitskiy2021} & 344.06         & 14.54         \\
DPT~\cite{Ranftl2021}     & BEiT-L~\cite{Bao2022}        & 345.01         & 9.60          \\
NeWCRFs~\cite{Yuan2021}   & Swin~\cite{Liu2021b}         & 270.44         & 21.61         \\
\bottomrule
\end{tabular}

%% file: Tables/res_abl.tex
% Ablation experiments. 

\addtolength{\tabcolsep}{-0.1em}

\mycaption{Leave-one-out Ablation}{%
We study the contribution of each proposed component. 
Randomizing the support frames, learning camera parameters and augmenting the image shape all contribute to improving overall performance. 
}

\begin{adjustbox}{center}
\begin{tabular}{@{}llllllllllll@{}}
\toprule
                            & \multicolumn{2}{c}{\textbf{Multi-task}} &           \multicolumn{3}{c}{\textbf{\acl{kez}}} &             \multicolumn{3}{c}{\textbf{\acs{mc}}} & \multicolumn{3}{c}{\zs\textbf{\acs{syns} (Val)}} \\ 
                                                            \cmidrule(lr){2-3}                                  \cmidrule(lr){4-6}                                \cmidrule(lr){7-9} \cmidrule(lr){10-12}
                            & R\down & \acs{relimp}\up &  Rel\down &          F\up & \acs{d25}\up &      Rel\down &          F\up & \acs{d25}\up &      Rel\down &          F\up & \acs{d25}\up \\ 
\midrule
  \textbf{Full} &  \best{2.20} &   \best{0.00} & \nbest{6.16} & \nbest{57.60} &     \nbest{95.52} &         14.39 &         17.67 &     \nbest{82.23} & \nbest{20.34} &         17.08 &      \best{69.88} \\ 
\midrule
                  Fwd \acs{Pose} &         2.60 & \nbest{-0.08} &         6.18 &         57.47 &             95.47 &         14.36 &         17.52 &             82.22 &  \best{20.24} &  \best{17.20} &             69.52 \\ 
$\acs{offset} = \pm1$ & \nbest{2.30} &         -0.60 &  \best{6.03} &  \best{58.23} &      \best{95.67} &  \best{14.17} &  \best{17.92} &      \best{82.43} &         21.04 &         16.04 &             68.33 \\ 
            Fixed \acs{Cam} &         4.00 &         -1.46 &         6.30 &         56.93 &             95.38 &         14.95 &         17.11 &             81.00 &         20.46 & \nbest{17.11} &     \nbest{69.56} \\ 
              No \ac{araug} &         4.50 &         -4.72 &         7.42 &         52.89 &             93.99 & \nbest{14.32} & \nbest{17.87} &             82.16 &         21.32 &         16.10 &             67.64 \\ 
                       None &         5.40 &         -5.52 &         7.47 &         51.83 &             94.19 &         14.62 &         17.01 &             81.29 &         21.21 &         16.72 &             67.03 \\ 
\bottomrule
\end{tabular}
\end{adjustbox}

\emph{%
Highlighted cells indicate \emph{\textbf{zero-shot}} results.
}

%% file: Figures/mdec.tex
% MDEC-2 test leaderoard

\includegraphics[width=1\linewidth]{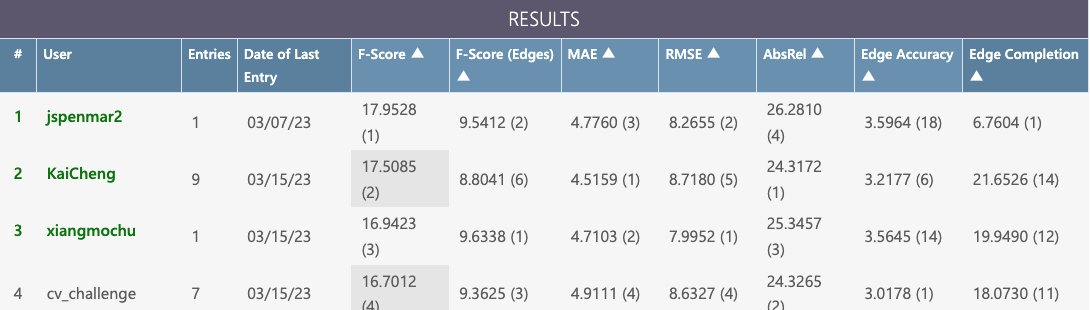}

\mycaption{MDEC-2~\cite{Spencer2023b}}{%
Our submission (\emph{jspenmar2}) was top of the MDEC-2 leaderboard in F-Score reconstruction. 
The challenge evaluated zero-shot performance on \acl{syns} for both supervised and self-supervised approaches.
}

%% file: Tables/res_reloc.tex
% MapFreeReloc benchmark results.

\mycaption{Map-free Relocalization~\cite{Arnold2022}}{%
We incorporate \acs{kbr} into a feature-matching pipeline for singe-image relocalization. 
We once again outperform the \acl{ssmde} baselines in every metric and perform on par with supervised \ac{sota}.
}

\begin{tabular}{@{}lllllllll@{}}
\toprule
                              &       &                                    \multicolumn{4}{c}{Pose} &                       \multicolumn{3}{c}{VCRE} \\
                                                                                 \cmidrule(lr){3-6}                               \cmidrule(lr){7-9}
                              & Train &   Trans\down &      Rot\down &         P\up &        AUC\up &     Error\down &          P\up &        AUC\up \\
\midrule
         Garg~\cite{Garg2016} &     S &         2.96 & \nbest{52.57} &         5.43 &         17.15 &         188.20 &         24.84 & \nbest{51.61} \\ 
 Monodepth2~\cite{Godard2017} &    MS &         2.95 &         52.92 &         5.50 &         17.22 &         189.67 &         24.38 &         50.63 \\ 
      DiffNet~\cite{Zhou2021} &    MS &         2.97 &         53.19 &         5.65 &         17.71 &         188.80 &         24.78 &         51.24 \\ 
      HR-Depth~\cite{Lyu2021} &    MS & \nbest{2.94} &         52.95 & \nbest{5.67} & \nbest{17.95} & \nbest{187.83} & \nbest{25.06} &         51.52 \\ 
\midrule
    \textbf{\acs{kbr} (Ours)} &     M &  \best{2.63} &  \best{49.01} & \best{11.54} &  \best{32.02} &  \best{181.21} &  \best{29.96} &  \best{58.89} \\ 
\midrule
\midrule
      MiDaS~\cite{Ranftl2020} &     D &         2.60 &         46.92 & \nbest{11.39} &         30.44 & \nbest{180.64} &         30.45 &         59.72 \\ 
    DPT-ViT~\cite{Ranftl2021} &     D & \nbest{2.56} & \nbest{45.62} &         11.27 & \nbest{30.92} &         181.34 & \nbest{30.60} & \nbest{60.03} \\ 
   DPT-BEiT~\cite{Ranftl2021} &     D &  \best{2.49} &  \best{44.99} &  \best{12.56} &  \best{32.48} &         181.67 &  \best{32.46} &  \best{62.03} \\ 
    NeWCRFs~\cite{Weihao2022} &     D &         2.89 &         51.92 &          6.69 &         20.77 &         184.63 &         25.89 &         52.93 \\ 
\midrule 
   DPT-NYUD~\cite{Arnold2022} &  D+FT &         2.67 &         47.66 &          9.17 &         26.46 &         184.53 &         28.68 &         56.87 \\ 
  DPT-Kitti~\cite{Arnold2022} &  D+FT &         2.66 &         49.21 &         10.86 &         29.99 &  \best{178.49} &         28.37 &         56.86 \\ 
\bottomrule
\end{tabular}

\emph{%
\textbf{Trans}=meters, \textbf{Rot}=deg, \textbf{VCRE}=px, \textbf{Precision}=\%, \textbf{AUC}=\%.
}

%% file: appendices.tex
\appendix

%-------------------------------------------------------------------------
\section{\acl{stv} Dataset} \label{sec:supp:data}
\fig{supp:data:slowtv} shows a frame from each \acl{stv} video, while \fig{supp:data:slowtv_map} shows their map location.
Sequences [00-27] are hiking scenes, [28-30] scuba diving and [31-39] driving. 
As seen, this dataset provides an incredible diversity of environments and locations, enabling us to train models capable of generalizing to previously unseen scene types.

%-------------------------------------------------------------------------
\section{Aspect Ratio Augmentation} \label{sec:supp:aug}
To make the models invariant to the training image size, we propose to incorporate an aspect ratio augmentation.
For more information see \sct{meth:aug} in the main paper.
Sample training images obtained using this procedure an be found in \fig{supp:aug}.
The centre crop is uniformly sampled from a set of predetermined aspect ratios:
% =====================================
\begin{itemize}
    \item Portrait: 6:13, 9:16, 3:5, 2:3, 4:5, 1:1
    \item Landscape: 5:4, 4:3, 3:2, 14:9, 5:3, 16:9, 2:1, 24:10, 33:10, 18:5
\end{itemize}
% =====================================

% =====================================
%
\begin{figure}[t]
\centering
\input{Figures/supp_araug}
\label{fig:supp:aug}
\end{figure}

% =====================================

%-------------------------------------------------------------------------
\section{Evaluation Datasets} \label{sec:supp:eval_data}
\heading{\acl{keb}~\cite{Geiger2013}} (Test: 652)
Subset of the common \acl{ke} split with corrected \acs{lidar}~\cite{Uhrig2018}.  

\heading{\acl{kez}~\cite{Geiger2013}} (Val: 700)
Subset of the \acl{kez} val split with corrected \acs{lidar}~\cite{Uhrig2018}.  

\heading{\acl{mc}~\cite{Geiger2013}} (Test: 1k)
Subset of the original test split, using COLMAP~\cite{Schoenberger2016} depth reconstructions. 

\heading{\acl{syns}~\cite{Adams2016,Spencer2022}} (Val: 400, Test: 775)
Official val and test splits consisting of dense \acs{lidar} maps. 

\heading{\acl{ddad}~\cite{Guizilini2020}} (Test: 1k)
Subset of the official val split, featuring \acs{lidar} maps with an increased range up to 250m. 

\heading{\acl{sintel}~\cite{Geiger2013}} (Test: 1064)
Official test split, consisting of synthetic image \& depth pairs from highly dynamic scenes

\heading{\acl{dii}~\cite{Vasiljevic2019}} (Test: 325)
Official val split with dense \acs{lidar} depth maps.

\heading{\acl{dio}~\cite{Vasiljevic2019}} (Test: 446)
Official val split with dense \acs{lidar} depth maps.

\heading{\acl{nyud}~\cite{Silberman2012}} (Test: 654)
Official test split collected using a Kinect RGB-D camera.

\heading{\acl{tum}~\cite{Geiger2013}} (Test: 2.5k)
Subset of dynamic scenes with moving people also collected using a Kinect.

%-------------------------------------------------------------------------
\section{Leaning Camera Intrinsics} \label{sec:supp:learnk}
Estimating the intrinsics parameters is required when training with uncalibrated cameras. 
However, this procedure can be applied even if the camera parameters are known. 
\tbl{supp:learnk} shows results when training on either \acl{keb} or \acl{mc}. 
If the dataset provides accurately calibrated cameras (Kitti), self-supervised learning of the intrinsics is on par with using the ground-truth parameters. 
However, when the ground-truth parameters are estimated using COLMAP~\cite{Schoenberger2016}, learning the intrinsics can slightly improve performance. 

% =====================================
%
\begin{table}[t]
\scriptsize
\addtolength{\tabcolsep}{-0.6em}
\renewcommand{\arraystretch}{1.2}
\centering
\input{Tables/supp_learnk}
\label{tbl:supp:learnk}
\end{table}

% =====================================

% =====================================
%
\begin{figure}[htbp]
\centering
\input{Figures/*}
\label{fig:[}
\end{figure}
t]{supp_slowtv}{supp:data:slowtv}
% =====================================

% =====================================
%
\begin{figure}[t]
\centering
\input{Figures/supp_slowtv_map}
\label{fig:supp:data:slowtv_map}
\end{figure}

% =====================================

%-------------------------------------------------------------------------
\section{Dynamic Objects} \label{sec:supp:motion}
\acs{mde} models trained exclusively using monocular supervision are prone to artefacts from dynamic objects. 
For instance, vehicles moving at similar speeds to the camera can produce holes of infinite depth due to their static appearance across images. 
Meanwhile, other dynamic objects can result in underestimated depth when moving towards the camera, or overestimated depth when moving away from it. 
This is due to the additional motion causing incorrect correspondences in the warping procedure. 

% The proposed \acs{kbr} does not explicitly include constraints that address these dynamic objects. 
Existing approaches that address these dynamic objects~\cite{Gordon2019,Casser2019,Dai2020} rely on additional labels such as semantic or instance segmentation. 
We instead opt for the losses proposed by Monodepth2~\cite{Godard2019} as a simpler proxy without increased computation or label requirements.

% =====================================
%
\begin{table}[htbp]
\scriptsize
\addtolength{\tabcolsep}{-0.6em}
\renewcommand{\arraystretch}{1.2}
\centering
\input{Tables/*}
\label{tbl:[}
\end{table}
t]{supp_motion}{supp:motion}
% =====================================

% =====================================
%
\begin{figure}[htbp]
\centering
\input{Figures/*}
\label{fig:[}
\end{figure}
t]{supp_preds_motion}{supp:motion}
% =====================================

We test the effectiveness of these constraints on a smaller subset of all three training datasets.
These results can be found in \tbl{supp:motion} and \fig{supp:motion}. 
Despite not explicitly modelling dynamic objects, Monodepth2 drastically increases the accuracy and robustness. 
This can be seen both in the improved metrics and the reduction in visual artefacts.

%-------------------------------------------------------------------------
\section{Median Alignment Results} \label{sec:supp:res}
\tbl{supp:res:all} shows results when applying median depth alignment between prediction and ground-truth. 
As expected, this generally results in worse performance that estimating both scale and shift parameters.
This is particularly noticeable for \acl{midas}, DPT and the \acs{ssl} baselines.

% =====================================
%
\begin{table}[htbp]
\scriptsize
\addtolength{\tabcolsep}{-0.6em}
\renewcommand{\arraystretch}{1.2}
\centering
\input{Tables/*}
\label{tbl:[}
\end{table}
t]{supp_res}{supp:res:all}
% =====================================

% =====================================
%
\begin{figure}[htbp]
\centering
\input{Figures/*}
\label{fig:[}
\end{figure}
t]{supp_preds_fail}{res:viz:fail}
% =====================================

%-------------------------------------------------------------------------
\section{Failure Cases} \label{sec:supp:fail}
Whilst representing a significant milestone in \acl{ssmde}, our model still suffers from several failure cases. 
We show these in \fig{res:viz:fail}.
For instance, Kitti shows a car estimated as a hole of infinite depth, despite training with the minimum reconstruction loss and automasking~\cite{Godard2019}.
Several visualizations are also characterized by texture-copy artefacts.
In some cases, our models estimated incorrect relative object positions (\eg \acs{sintel} or \acl{ddad}).
An interesting failure case for all approaches are highly-reflective surfaces, such as mirrors or TVs.
These are challenging due to the fact that they do not violate the photometric error and obtaining \acs{lidar} or \ac{sfm} ground-truth is highly challenging. 
Finally, due to the strong prior for upright images, our model struggles to adapt to extreme rotations (\acl{tum}). 
This could be mitigated with additional augmentations. 
Finally, it is worth pointing out that, in the vast majority of these cases, our model outperforms the \acs{ssl} baselines. 

\clearpage

%% file: Figures/supp_araug.tex
% Supplementary aspect ratio augmentation visualizations.

\vspace{-0.25cm}

\subfloat[Original (16:9)]{\includegraphics[width=0.69\linewidth]{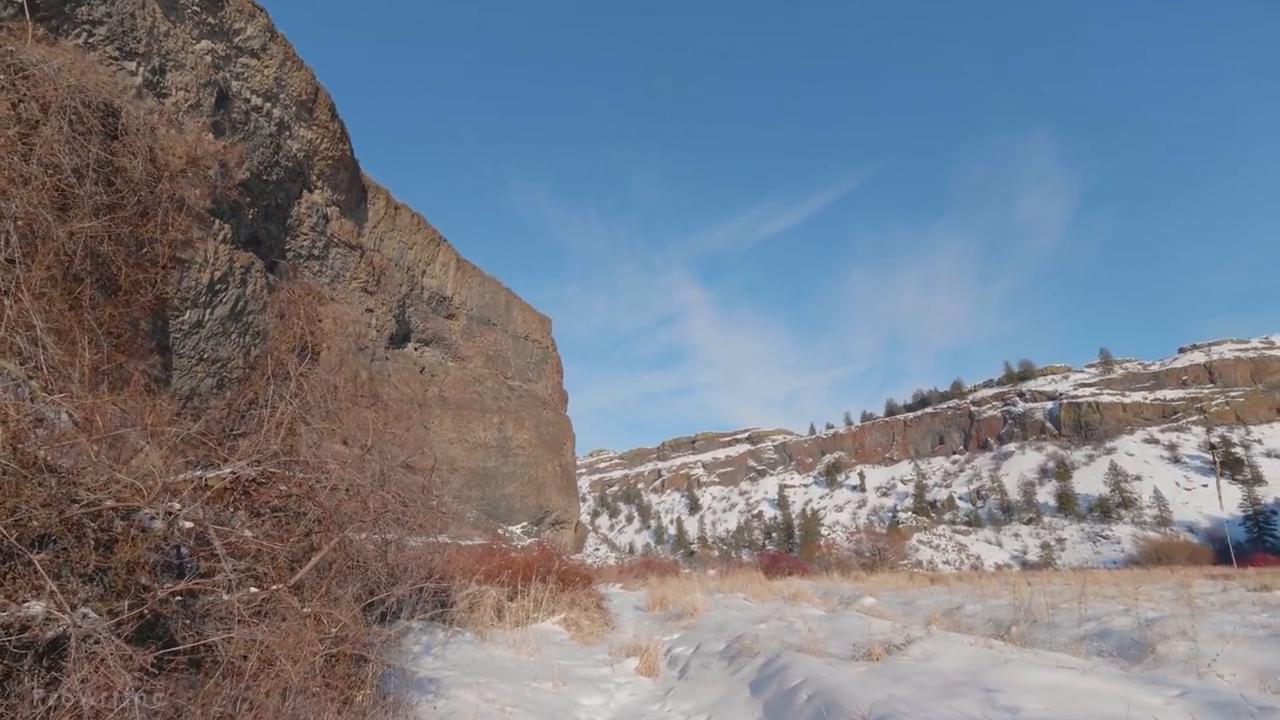}}\hfill
\subfloat[4:5]{\includegraphics[width=.29\linewidth]{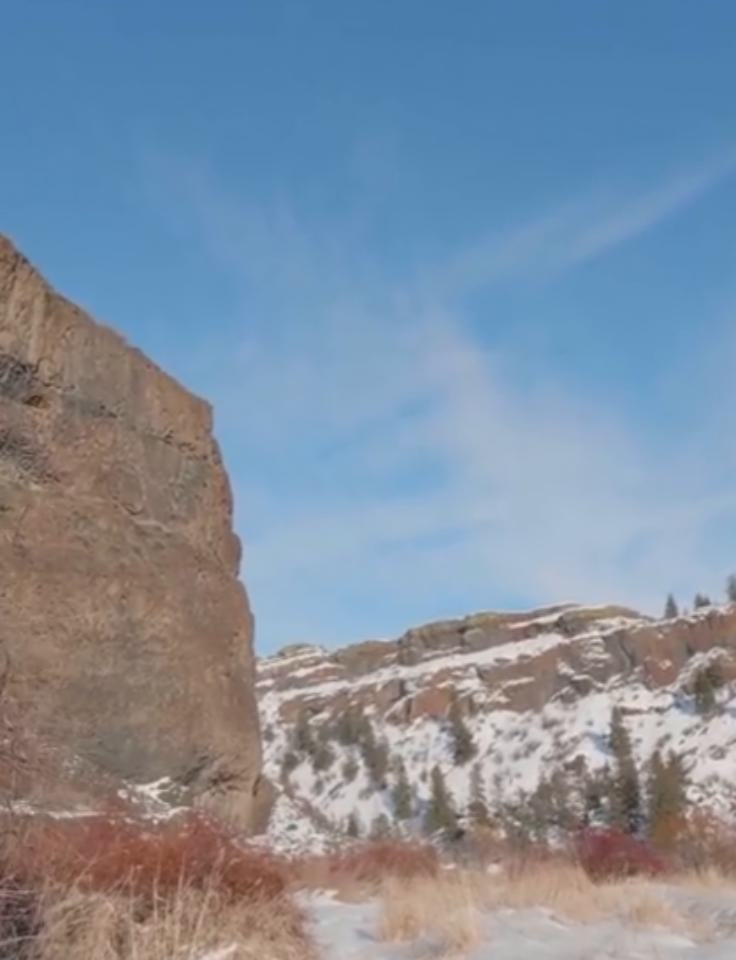}}\hfill
\subfloat[Original (16:9)]{\includegraphics[width=0.49\linewidth]{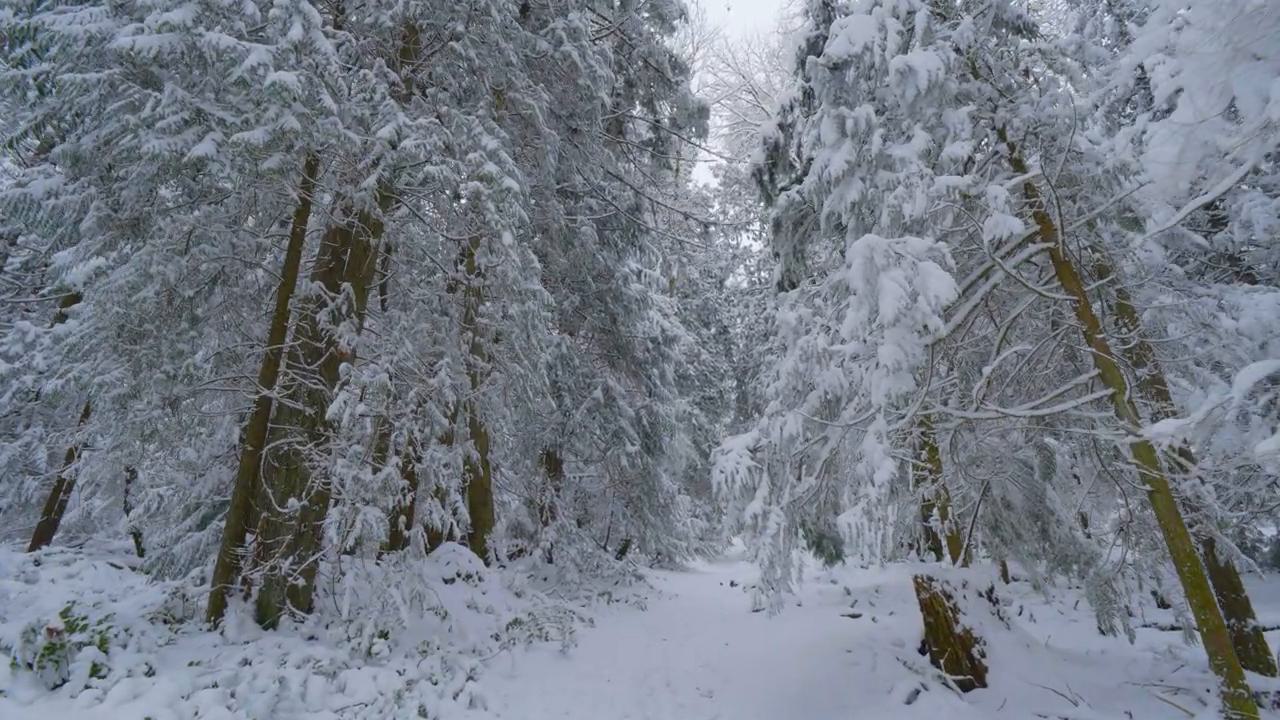}}\hfill
\subfloat[5:3]{\includegraphics[width=.49\linewidth]{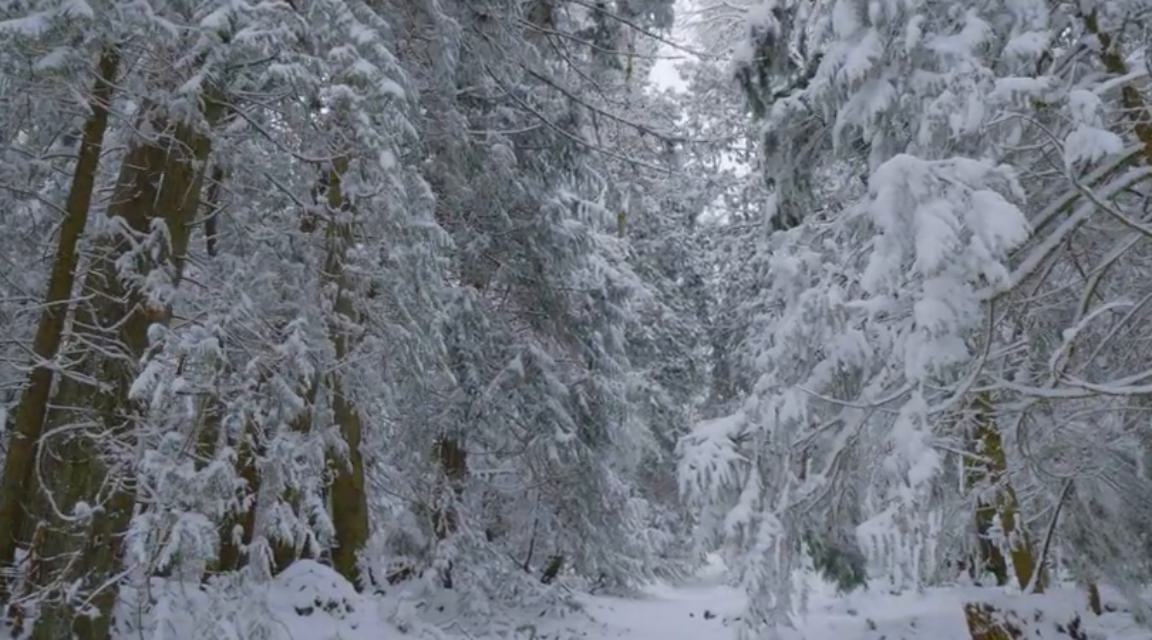}}
\\[-0.5ex]
\subfloat[Original (16:9)]{\includegraphics[width=0.49\linewidth]{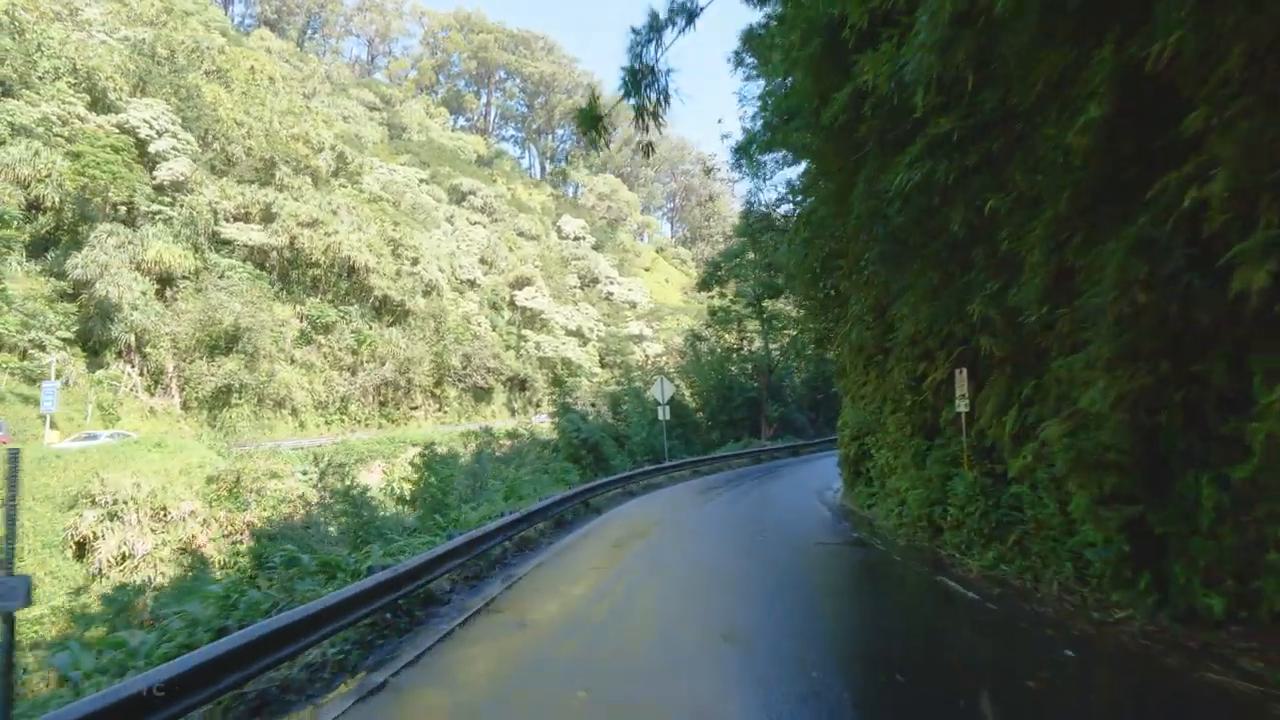}}\hfill
\subfloat[2:1]{\includegraphics[width=.49\linewidth]{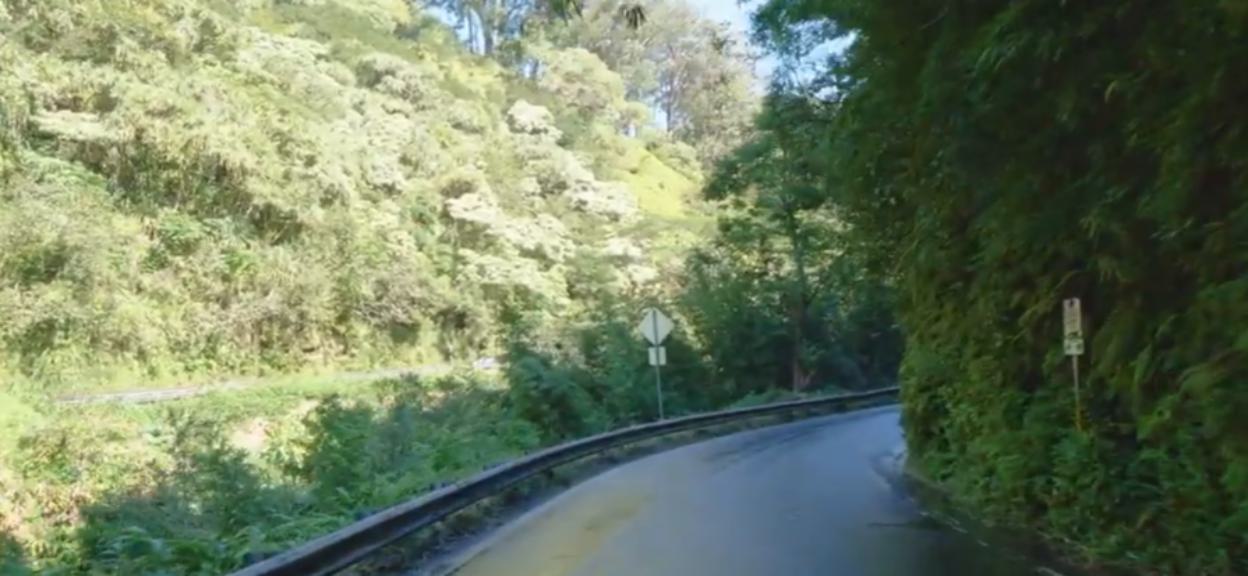}}\hfill
\subfloat[Original (16:9)]{\includegraphics[width=0.59\linewidth]{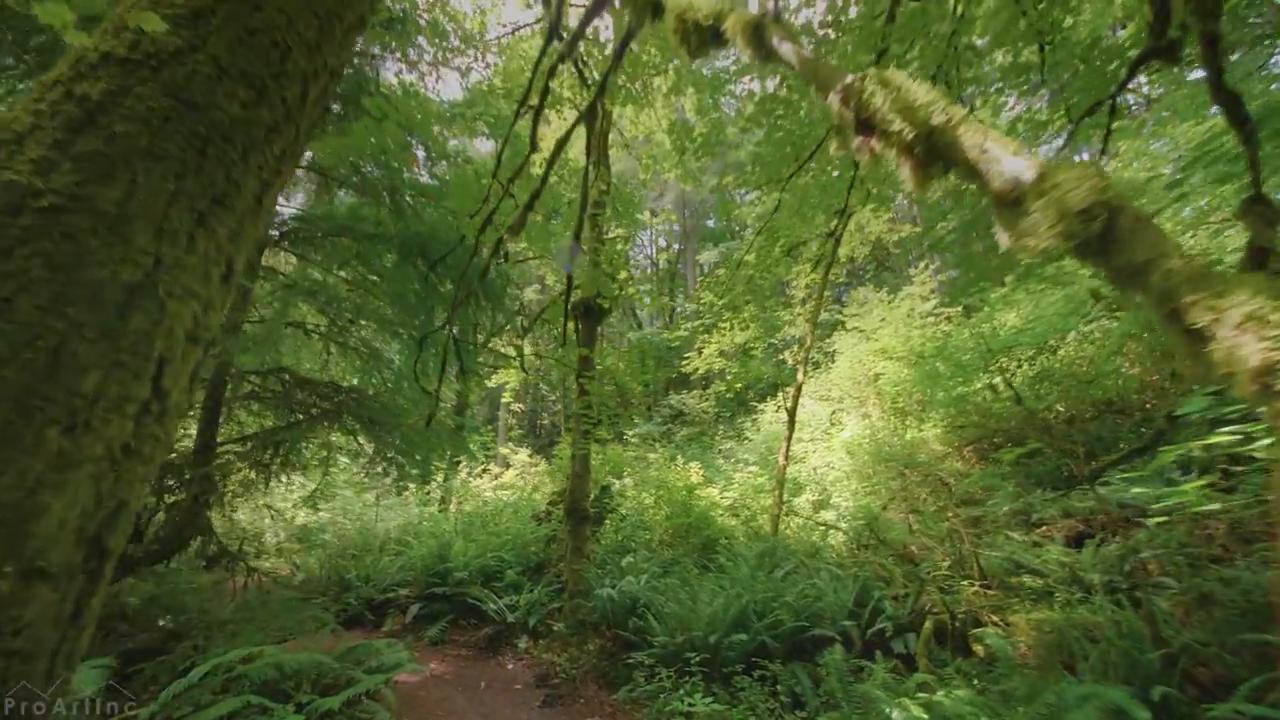}}\hfill
\subfloat[1:1]{\includegraphics[width=.39\linewidth]{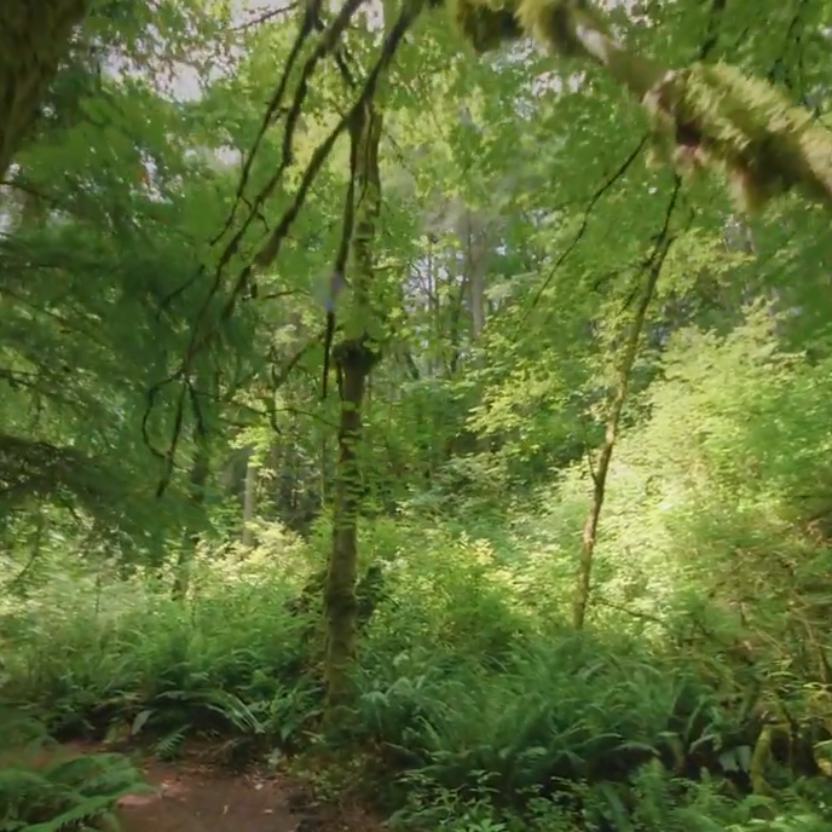}}

\mycaption{\acs{araug}}{%
Additional augmentations used to diversify the variety of image shapes and object scales seen by the network.
}
\vspace{-0.5cm}

%% file: Tables/supp_learnk.tex
\footnotesize

\mycaption{Learning Camera Intrinsics}{%
Performance when training on a single dataset (Kitti or \acl{mc}) and learning camera intrinsics. 
If the cameras are not perfectly calibrated, learning the intrinsics can improve accuracy. 
}

\begin{tabular}{@{}lllllll@{}}
\toprule
              &    \multicolumn{3}{c}{\textbf{\acl{kez}}} \\ 
              \cmidrule(lr){2-4}  
              &     Rel\down &          F\up & $\delta_{.25}$\up    \\ 
\midrule
     Baseline  & \nbest{5.69} &  \best{60.88} &     \nbest{95.89}  \\ 
Learn \acs{Cam} &  \best{5.68} & \nbest{60.81} &      \best{95.90} \\ 
\bottomrule
\end{tabular}
\quad
\begin{tabular}{@{}lllllll@{}}
\toprule
              &     \multicolumn{3}{c}{\textbf{\acs{mc}}} \\ 
              \cmidrule(lr){2-4}  
                &      Rel\down &          F\up & $\delta_{.25}$\up \\ 
\midrule
     Baseline   & \nbest{16.66} & \nbest{14.20} &     \nbest{77.18} \\ 
Learn \acs{Cam} &  \best{16.12} &  \best{14.77} &      \best{78.40} \\ 
\bottomrule
\end{tabular}

%% file: Figures/supp_slowtv_map.tex
% Supplementary SlowTV video locations on world map.

\includegraphics[width=1\linewidth]{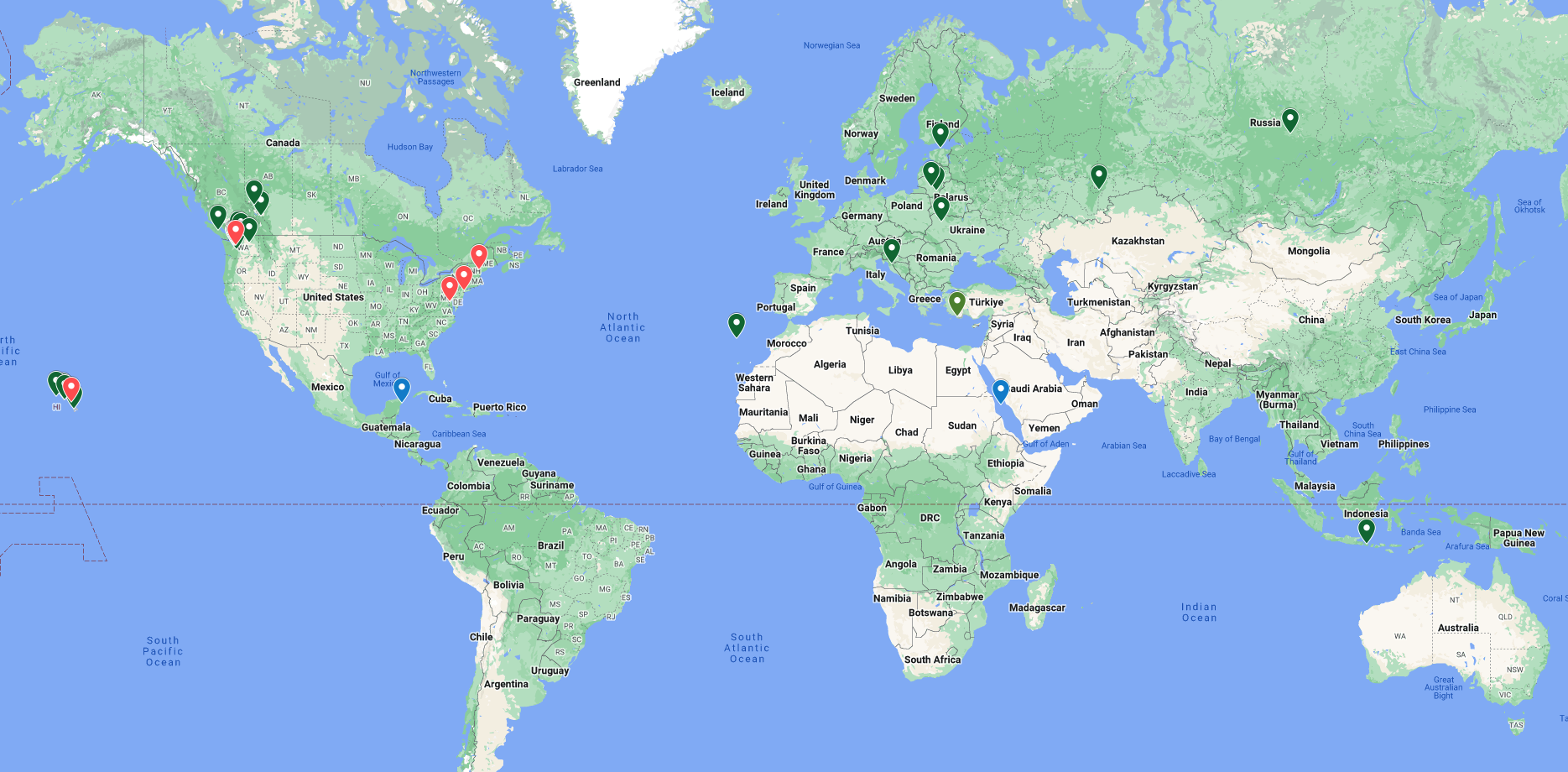}

\mycaption{\acl{stv} Map}{%
Distribution of locations in the proposed dataset.
\textcolor{ForestGreen}{\textbf{Green}}=Natural, \textcolor{red}{\textbf{Red}}=Driving, \textcolor{blue}{\textbf{Blue}}=Underwater.
}